\documentclass[11pt]{article}

\usepackage[margin=1in]{geometry}
\usepackage{times}
\usepackage{amsmath,amssymb}
\usepackage{booktabs}
\usepackage{array}
\usepackage{graphicx}
\usepackage{url}
\usepackage[hidelinks]{hyperref}
\usepackage{xcolor}
\usepackage{caption}
\usepackage{authblk}
\usepackage{microtype}
\usepackage{enumitem}
\usepackage{tikz}
\usetikzlibrary{arrows.meta,positioning,calc,fit,backgrounds}
\usepackage{pgfplots}
\pgfplotsset{compat=1.16}
\usepgfplotslibrary{groupplots}
\usepackage{tcolorbox}
\setlist{itemsep=2pt,topsep=3pt}

\newcommand{\best}[1]{\textbf{#1}}
\newcommand{\EWT}{\textsc{EW-T}}
\newcommand{\EWA}{\textsc{EW-A}}
\newcommand{\EWB}{\textsc{EW-B}}

\definecolor{cfrozen}{RGB}{225,225,225}
\definecolor{ctrained}{RGB}{200,220,245}
\definecolor{cfixed}{RGB}{205,235,205}
\definecolor{ctext}{RGB}{255,240,190}
\definecolor{cbad}{RGB}{250,215,210}
\definecolor{cgood}{RGB}{190,235,200}

\newtcolorbox{findings}[1][Findings]{colback=blue!4!white,colframe=blue!45!black,boxrule=0.5pt,arc=1.5pt,
  left=6pt,right=6pt,top=3pt,bottom=3pt,fontupper=\small,title=#1,fonttitle=\small\bfseries,
  before skip=8pt,after skip=10pt}

\title{\bf From LoRA to Language:\\
Recovering Skill Documents from Adapters}

\author{Saranya Dintakurthy Swadha}
\author{Pavan C Shekar}
\author{Arpit Jain}
\author{Aswanth Krishnan}
\affil{QpiAI, Bengaluru, India}
\affil{\normalsize\texttt{\{saranya.d, pavan.s, arpit.j, ashwanth.krishnan\}@qpiai.tech}}
\date{}

\begin{document}
\maketitle

\begin{abstract}
Agent skills are short documents that teach a language model how to handle a class of tasks. To save context, these documents are
increasingly compiled into low-rank adapters (LoRA), which keep the behaviour but lose the text: an adapter is hard to read, edit, audit or
move to another model. We ask whether the skill document can be written back from the adapter and its frozen base model, and we score each
recovered document by the share of the original's key instructions it states and by the tasks it solves. Reading the weights alone recovered
little with the methods we tried; a learned map from weights to text passed on about one bit of the document. Running the adapter worked much
better. The \emph{Executing Writer}, a small writer trained on top of the adapted model, recovers $39$--$44\%$ of the key instructions of
held-out skills, against $1.3$--$1.5\%$ with another skill's adapter, and on adapters fitted to $41$ external skills its documents restore
$85$--$91\%$ of the original's task gain. These strongest results are on adapters whose training stored the document; adapters trained on
behaviour alone are much harder for trained writers ($1$--$2\%$ of instructions). For these adapters
we test an agent that runs the base and adapted models and writes a skill from where they differ. On ALFWorld its skills recover $58$--$71\%$
of a LoRA's gain and carry over to models that cannot load the LoRA, lifting Llama-3.1-8B from $5\%$ to $55\%$. Whether such a skill comes from
the adapter depends on what the adapter learned: when its behaviour cannot be guessed, the adapter is the source; when it can, a reward loop
without the adapter does as well.
\end{abstract}


\section{Introduction}

An \emph{agent skill} is a document, usually a few hundred to a few thousand words, that tells a language model how to handle a class of
requests: how to review code, which conventions to follow, how to analyse a packet capture~\cite{anthropicskills,skillsbench}. Skills are
written by practitioners, shared in marketplaces and placed in the model's prompt when needed. That costs context on every call, so a growing
line of work compiles a skill into a \emph{low-rank adapter} (LoRA)~\cite{lora}, a small set of extra weights that changes the model's behaviour
with no text in the prompt~\cite{t2l,d2l,shine,latentskill,paramskills}. The adapter is cheap to apply and to serve at scale~\cite{slora}, but it
is a block of numbers. A text skill can be read before it is trusted; a weight skill is much harder to check.

\paragraph{The question.} We study the inverse of compilation:
\begin{quote}\itshape
Given an adapter and its frozen base model, can we write the skill document that the adapter encodes?
\end{quote}
A recovered document would make an adapter inspectable and editable, could be re-compiled for another model, and would let a user verify an
adapter before running it. Existing weight-space methods predict attributes, labels or short descriptions of adapters~\cite{w2t,loracles,dit,weightwatch,spectral};
none of them produces a document or measures how much of one is recovered.

\paragraph{Why a document and not only an adapter.} A skill document and a LoRA can carry the same behaviour, but they are not equally useful.
A document can be read and reviewed like code, edited in one line, combined with other skills, and given unchanged to a different model or
agent harness. A LoRA is tied to the weights and tokenizer of the base model it was trained on: it does not load on Llama if it was trained
on Qwen, nor on a model served only through an API, and changing one rule means retraining it. Recovering the document turns a model-locked
artefact back into a portable one. We therefore measure recovery in two ways: the share of the original's instructions that the recovered
document states, and the share of the original's task gain that it restores.

\paragraph{Why it is hard.} Three properties of adapters shape the methods. (i) Adapters for different skills are mostly alike: the cosine
between the weight updates of two skills is $0.88$, and only $11.6\%$ of their energy is specific to the skill. (ii) Knowing which skill an
adapter holds is not knowing what it says. An encoder identifies the skill from its weights for all $37$ test skills, yet a learned channel from
the weights to a writer passes on about one bit of the document's content. (iii) Text similarity is a weak measure, because word overlap rewards generic text. We therefore score
recovered documents against frozen lists of the original's instructions, run every method with a wrong adapter as a control, and measure tasks
solved wherever we can.

\begin{figure}[t]
\centering
\includegraphics[width=0.78\linewidth]{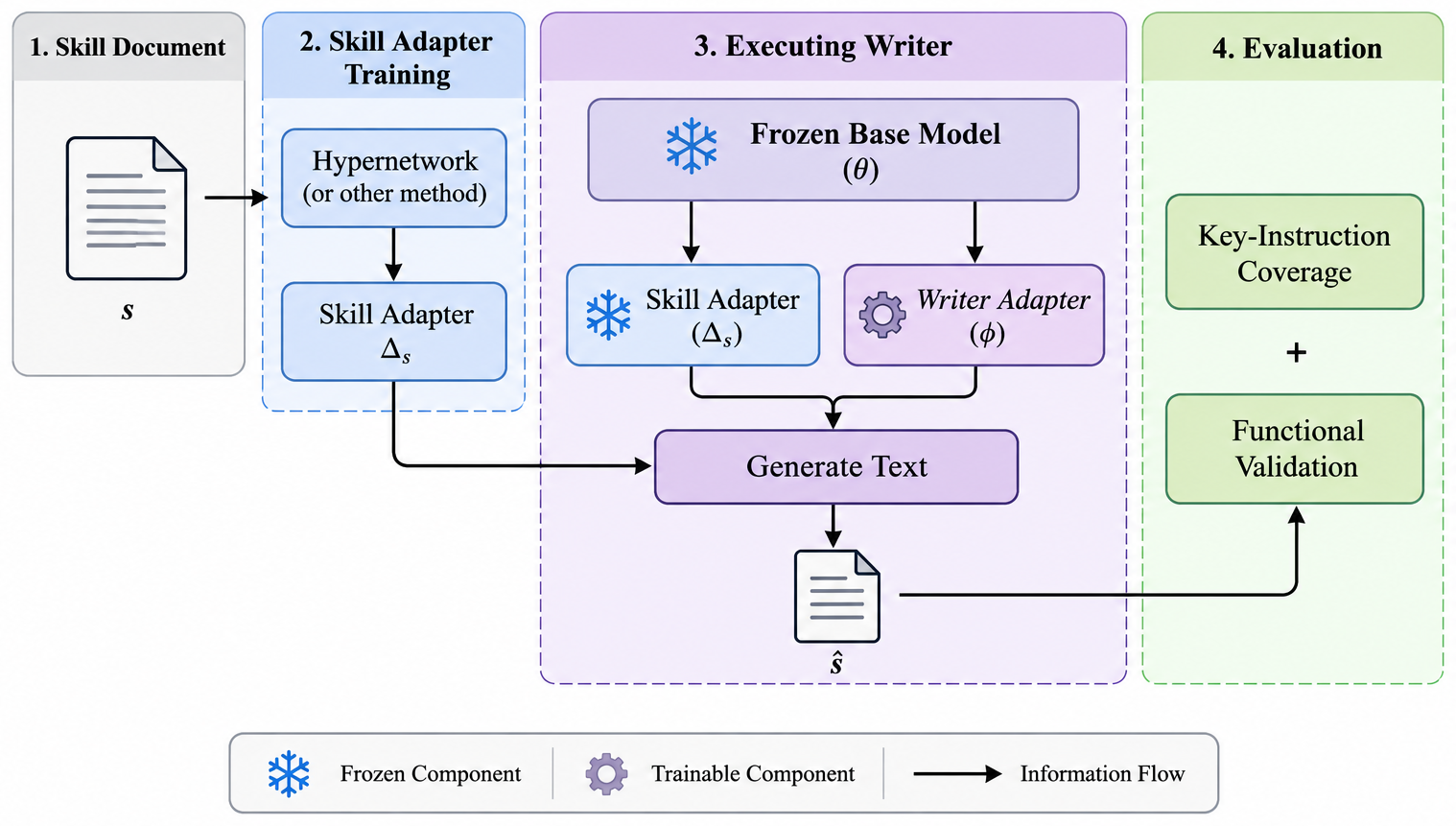}
\caption{The pipeline. A skill document $s$ is compiled into an adapter $\Delta_s$. The Executing Writer attaches $\Delta_s$ (frozen) and a
trainable writer $\phi$ to the frozen base model $\theta$ and generates a document $\hat s$. Recovery is scored by the share of the original's
key instructions that $\hat s$ states and by whether $\hat s$ still works when given to a model.}
\label{fig:pipeline}
\end{figure}

\paragraph{Our approach.} The content of an adapter shows in what the network computes with it, so we let the network do the reading.
We compare three readers (Figure~\ref{fig:pipeline}, Section~\ref{sec:method}). The first treats the weights as data and learns a map from
them to a prompt. The second reads the weights without running them, through the model's vocabulary and by retrieving similar instruction
lines. The third, the \emph{Executing Writer}, attaches the adapter to its frozen base model and trains one small writer adapter on top, so
that the adapted model itself writes the document. Section~\ref{sec:ax} adds a fourth reader, which needs no training and suits adapters that
never saw their document: an agent that runs the base and adapted models on real tasks and writes a skill from where they differ.

\paragraph{One distinction matters throughout.} Adapters differ in whether the document text was ever a training target. A \emph{recitation}
adapter was trained to print its document, so the text is stored in it, and recovering it is extraction. A \emph{behavioural} adapter was
trained only on answers that the document shaped, so the text never appears in any training target, and recovering it is inference. The second
case is the common one in practice. Our trained writers read the first kind well and the second poorly; for the second kind we turn to the
executing agent.

\paragraph{Contributions.}
\begin{enumerate}[leftmargin=*]
\item We define adapter-to-document recovery, with coverage of frozen instruction lists, a measure of invented content, and two controls
the evidence needs: a wrong adapter and retrieval of a similar document.
\item We compare three readers (Sections~\ref{sec:method} and~\ref{sec:res-doc}). The weights reveal a skill's identity and content words,
but a learned soft-prompt channel passes on about one bit; an information measure, the \emph{adapter gap}, shows where the difference lies.
\item The Executing Writer recovers $39$--$44\%$ of a held-out skill's key instructions, about twice the best weight-only writer and $1.4$--$1.6$ times retrieval, and only
$1.3$--$1.5\%$ with a wrong adapter. On adapters fitted to $41$ external skills it recovers $90$--$93\%$ of the prose at rank $64$; these
documents restore $85$--$91\%$ of the original's task gain and help an agent on benchmark tasks almost as much as the original.
\item For behavioural adapters (Section~\ref{sec:ax}) we test an executing agent on ALFWorld, a docker-compose adapter from an earlier project, synthetic tasks and
adapters with hidden rules. Its skills recover $58$--$71\%$ of an ALFWorld LoRA's gain and transfer to other models. Controls show that the
adapter is the source of the skill only when the adapter's behaviour cannot be guessed.
\end{enumerate}

\section{Related Work}

\paragraph{Agent skills.} Skills package procedures and conventions as documents that an agent loads when a task calls for
them~\cite{anthropicskills}. SkillsBench measures how much skills help on diverse tasks with independent verifiers~\cite{skillsbench}, and
earlier agents kept libraries of learned skills~\cite{voyager} or interleaved reasoning with tool use~\cite{react,sweagent}. We use
SkillsBench skills and tasks, which we did not write, to test whether recovered documents still work.

\paragraph{Compiling text into weights.} LoRA~\cite{lora} represents a weight update by low-rank factors, following adapters~\cite{houlsby}
and continuous prompts~\cite{prefix}, with evidence that fine-tuning moves a model in a low-dimensional subspace~\cite{intrinsic} and that LoRA
learns less than full fine-tuning~\cite{lorabiderman}. Hypernetworks~\cite{hypernet} generate weights from other inputs: HYPTER~\cite{hypter},
Text-to-LoRA~\cite{t2l}, Doc-to-LoRA~\cite{d2l} and SHINE~\cite{shine} generate adapters from task descriptions or documents, and
LatentSkill~\cite{latentskill} and Parametric Skills~\cite{paramskills} do this for agent skills. Context distillation~\cite{askell,snell},
gist tokens~\cite{gist} and cartridges~\cite{cartridges} also move context out of the prompt, and adapters can be composed~\cite{lorahub} or
added arithmetically~\cite{taskarith}. All of these go from text to weights; we go the other way.

\paragraph{Prompt and skill optimisation.} Searching over prompts with a model in the loop is an established way to improve instructions:
APE~\cite{ape}, OPRO~\cite{opro}, DSPy~\cite{dspy}, TextGrad~\cite{textgrad} and GEPA~\cite{gepa} propose and score candidate prompts, and
Reflexion~\cite{reflexion}, ExpeL~\cite{expel} and Agent Workflow Memory~\cite{awm} turn an agent's experience into reusable text. Our reward
loop (Section~\ref{sec:ax}) belongs to this family; what is new is the adapter as an extra source of evidence, and the question of whether the
document it yields comes from that adapter. Work on auditing models for hidden objectives~\cite{auditing} asks a related question about behaviour
that a model was trained to hide.

\paragraph{Interpreting adapter weights.} W2T~\cite{w2t} and Learning on LoRAs~\cite{lol} process LoRA weights with symmetry-aware networks to
predict attributes, identity and performance, and hyper-representations embed~\cite{hyperrep} or generate~\cite{nndiff} weights. LoRAcles~\cite{loracles}
trains a model to answer questions about an adapter from its weights, Diff Interpretation Tuning~\cite{dit} trains a fine-tuned model to describe its
own change, WeightWatch~\cite{weightwatch} labels the top singular vectors of a weight difference, and spectral summaries identify the training
objective~\cite{spectral}. These produce labels, scores or short answers. We produce a document and measure how much of the original it contains.

\paragraph{Reading hidden states as text.} The logit lens~\cite{logitlens} and tuned lens~\cite{tunedlens} map a hidden direction to word
scores with the model's own output layer, and Patchscopes~\cite{patchscopes} and SelfIE~\cite{selfie} ask the model itself to describe a
hidden state. Models can describe behaviours they were trained on~\cite{betley}, and agents that experiment on a model can explain parts of
it~\cite{maia}. We apply the lens to an adapter's output directions after removing what adapters share, and we use the adapted model's own
generation, or an agent that runs it, as the reader.

\paragraph{Extraction and inversion.} Training data and prompts can be recovered from models and embeddings~\cite{carlini,nasr,vec2text,lminversion}.
Recovering a skill from an adapter inverts the compile step rather than attacking private data, but it raises the same question of what a
released artefact reveals (Section~\ref{sec:discussion}).

\section{Problem and Measures}
\label{sec:problem}

\paragraph{Adapter families.} What an adapter can reveal depends on how it was made, and Table~\ref{tab:families} lists the three families
we use. A recitation adapter exposes its text by construction, so results on recitation adapters are an upper bound on what a reader can do.
They do not reflect test leakage: every writer is scored on held-out skills it never saw.

\begin{table}[!htb]
\centering
\caption{The three adapter families used in this paper.}
\label{tab:families}
\small
\setlength{\tabcolsep}{4pt}
\begin{tabular}{>{\raggedright\arraybackslash}p{2.4cm}>{\raggedright\arraybackslash}p{7.6cm}>{\raggedright\arraybackslash}p{3.2cm}}
\toprule
Family & How it is made & Document in a training target?\\
\midrule
Document-bearing & A hypernetwork~\cite{latentskill} compiles the document into an adapter. One checkpoint is trained to store its
document (\emph{document-fitted}); a further-trained one is tuned for agent behaviour (\emph{task-tuned}). & yes for document-fitted, no for task-tuned\\
Recitation & We fit a LoRA by gradient descent to print the document after a trigger (ranks $16$, $32$, $64$). & yes, as the output\\
Behavioural & The model answers requests with the document in its prompt; a LoRA is fitted to give the same answers without it (context distillation). & no\\
\bottomrule
\end{tabular}
\end{table}

\paragraph{The recovery problem.} Let $\theta$ be a frozen base model and $s$ a skill document. An adapter $\Delta(s)=\{(A_b,B_b)\}_b$ adds a
low-rank update $\Delta W_b=A_bB_b$ to each weight block $b$, giving the adapted model $q_{\theta\oplus\Delta}(\cdot\mid x)$. \emph{Given $\Delta$
and $\theta$ but not $s$, produce a document $\hat s$ that states what $s$ states.}

\paragraph{Coverage.} For each original $s$ we extract once a frozen checklist of key instructions $C(s)=\{c_1,\dots,c_m\}$, so every method is
scored against the same list. A judge model~\cite{mtbench} assigns each $c_i$ one of \{full, partial, missing, contradicted\} with respect to $\hat s$, and
\[
\mathrm{Cov}(\hat s,s)=\frac1m\sum_{i=1}^{m}\Big(\mathbf 1[\text{full}_i]+\tfrac12\mathbf 1[\text{partial}_i]\Big).
\]
The primary judge returns a probability for each verdict, so coverage is an expectation. As a check, a judge from another model family scores every document again; Appendix~\ref{app:judge} compares the two judges. For code-heavy skills we score a \emph{prose} checklist (every instruction sentence) and a \emph{structure}
checklist (every code block, table and heading) separately, plus a judge-free count of identifiers such as paths, commands and flags.

\paragraph{Invented content.} Coverage ignores what a document adds. We split $\hat s$ into instruction units by a fixed rule and report the
share that the original does not support, $\mathrm{Unsup}(\hat s,s)$, as the \emph{unsupported} share.

\paragraph{Controls.} Every method is also run with another skill's adapter (\emph{wrong-adapter control}): a method that depends on the adapter
must fall to near zero. We also compare with \emph{retrieval} of the training document closest to the adapter, the natural way to use the weights
without writing anything.

\section{Method: Three Readers}
\label{sec:method}

\begin{figure}[!htb]
\centering
\begin{tikzpicture}[
  box/.style={draw, rounded corners=2pt, align=center, font=\footnotesize, minimum height=9mm, inner sep=3pt},
  frozen/.style={box, fill=cfrozen}, trained/.style={box, fill=ctrained}, fixed/.style={box, fill=cfixed},
  bad/.style={box, fill=cbad}, good/.style={box, fill=cgood}, arr/.style={-{Stealth[length=2.2mm]}, thick}]
\node[frozen, text width=2.2cm] (ad) at (0,0) {adapter $\Delta$\\{\scriptsize given, frozen}};
\node[trained, text width=2.6cm] (e1) at (4.4,2.5) {weight encoder\\{\scriptsize reads 252 blocks}};
\node[trained, text width=1.8cm] (w1) at (8.6,2.5) {writer};
\node[bad, text width=2.8cm] (o1) at (12.2,2.5) {$\approx$1 bit of content\\{\scriptsize ignores $\Delta$}};
\node[font=\footnotesize\bfseries, anchor=south west] at (3.1,3.05) {R1 \; weights as data};
\node[fixed, text width=2.6cm] (e2) at (4.4,0) {vocabulary lens\\+ retrieval head};
\node[trained, text width=1.8cm] (w2) at (8.6,0) {writer};
\node[bad, text width=2.8cm] (o2) at (12.2,0) {words, no order\\{\scriptsize 16--22\% of key instr.}};
\node[font=\footnotesize\bfseries, anchor=south west] at (3.1,0.55) {R2 \; vocabulary \& lines};
\node[frozen, text width=2.6cm] (e3) at (4.4,-2.5) {base model $\theta$\\ with $\Delta$ attached};
\node[trained, text width=1.8cm] (w3) at (8.6,-2.5) {writer $\phi$\\{\scriptsize rank 16}};
\node[good, text width=2.8cm] (o3) at (12.2,-2.5) {document $\hat s$\\{\scriptsize 39--44\% of key instr.}};
\node[font=\footnotesize\bfseries, anchor=south west] at (3.1,-1.95) {R3 \; execution (ours)};
\draw[arr] (ad.north) |- (e1.west); \draw[arr] (ad.east) -- (e2.west); \draw[arr] (ad.south) |- (e3.west);
\draw[arr] (e1) -- node[below,font=\scriptsize]{soft prompt} (w1); \draw[arr] (e2) -- node[below,font=\scriptsize]{text} (w2);
\draw[arr] (e3) -- node[below,font=\scriptsize]{runs} (w3);
\draw[arr] (w1) -- (o1); \draw[arr] (w2) -- (o2); \draw[arr] (w3) -- (o3);
\end{tikzpicture}
\caption{Three ways to read an adapter. R1 and R2 never run the adapter; R3 runs it inside its base model and trains only a small writer.
Grey: given and frozen; blue: trained; green: fixed computation. Recovery figures are from Table~\ref{tab:main}.}
\label{fig:overview}
\end{figure}
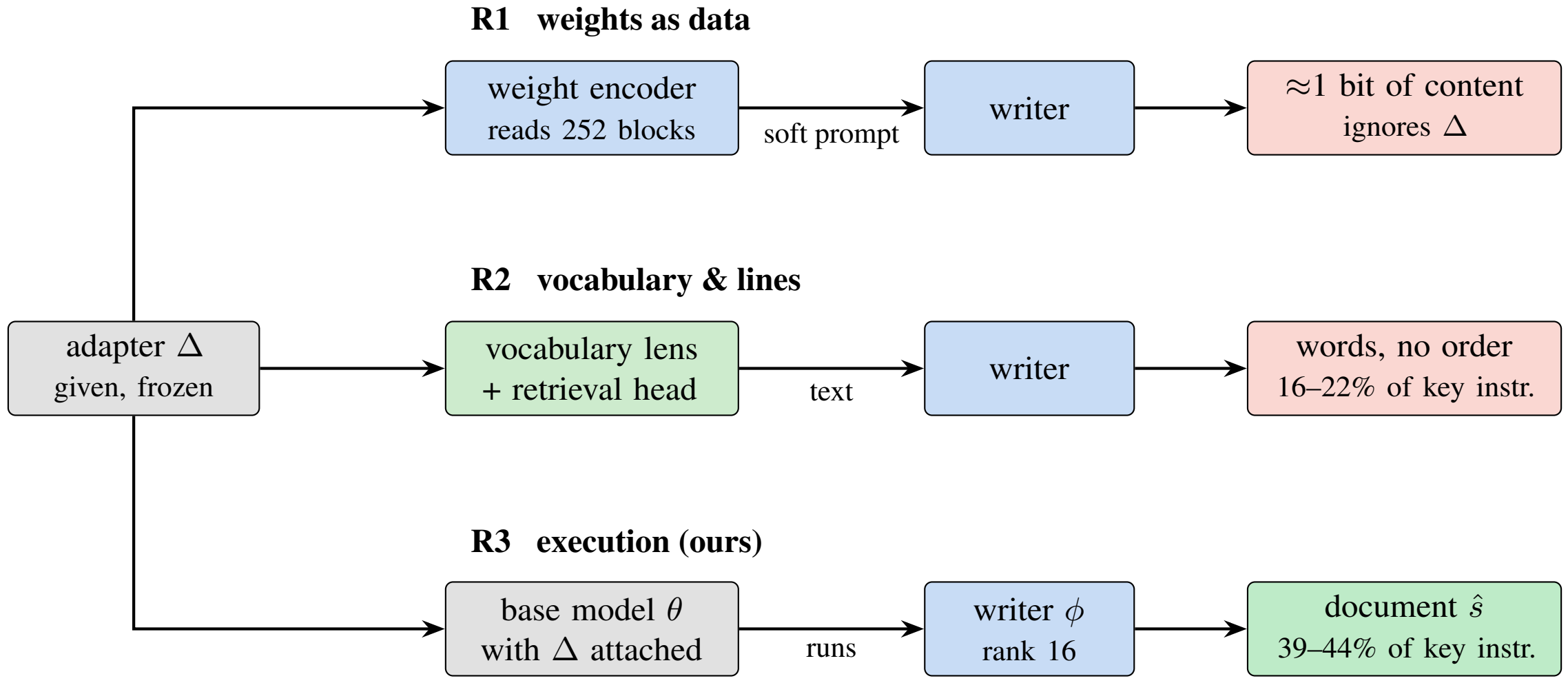

The three readers differ in what they ask a learned model to work out (Figure~\ref{fig:overview}). The first (R1) treats the adapter's numbers
as data: it encodes the complete adapter and trains a writer to produce the document from that encoding. The second (R2) looks up which words
the numbers point to and retrieves similar instruction lines, and a writer composes the document from this text. The third (R3, the Executing
Writer) does not read the numbers at all: it attaches the adapter to its frozen base model and lets the adapted model say what it knows. Only
R3 runs the adapter.

\paragraph{R1: the weights as data.} We encode the complete adapter (all $252$ blocks, no centring or projection) with a transformer into a short
learned soft prompt placed in front of a writer (Qwen3-4B or Qwen3-8B with a rank-$16$ adapter), which also receives the text that R2 reads.
Encoder and writer are trained jointly to produce the document. We tried five variants: a pooled vector, all block states with a margin loss on
the adapter gap (Section~\ref{sec:info}), a frozen decoder, vocabulary-aligned inputs, and the complete adapter with the 8B writer.

\paragraph{R2: the vocabulary and instruction retrieval.} The Gram matrix of $n$ adapters can be computed from the factors without forming $\Delta W$:
$G_{ij}=\sum_b\operatorname{tr}\big((A_b^{(i)\top}A_b^{(j)})(B_b^{(i)}B_b^{(j)\top})\big)$. Over $20$ test adapters the mean cosine between
different skills is $0.878$. Centring, $\tilde G=HGH$ with $H=I-\tfrac1n\mathbf 1\mathbf 1^{\top}$, leaves $11.6\%$ of the energy, and the centred
residuals have mean cosine $-0.051$, the value $-1/(n-1)$ expected for unrelated vectors. Each adapter is a large shared part plus a small,
distinct remainder. For each of the $72$ blocks that write into the residual stream we take the output directions $v_{b,k}$ of the centred update
and read them through the final norm $\gamma$ and output embedding, $z_{b,k}=W_U(\gamma\odot v_{b,k})$; the top $100$ words, weighted by
singular value, are the adapter's keywords. Of an adapter's top $40$ words, $73\%$ occur in its own document against $11\%$ in other test documents
($40\%$ against $8\%$ without centring). Words carry no order, so a \emph{retrieval encoder} ($4$ transformer layers, $7.5$M parameters) reads
compact linear summaries of the $72$ blocks and is trained with a contrastive loss~\cite{infonce,clip} against document embeddings~\cite{qwen3emb}
on $10{,}500$ training skills. It names the correct document for all $37$ test skills among $237$ candidates. An instruction head built on the same
encoder retrieves the $30$ closest lines from a bank of $208{,}995$ lines taken from training documents only; these lines match $24.6\%$ of the original's
instructions, against $0.2\%$ for a wrong adapter.

\begin{figure}[!htb]
\centering
\begin{tikzpicture}[
  box/.style={draw, rounded corners=2pt, align=center, font=\footnotesize, minimum height=8mm, inner sep=3pt},
  frozen/.style={box, fill=cfrozen}, trained/.style={box, fill=ctrained}, fixed/.style={box, fill=cfixed}, textn/.style={box, fill=ctext},
  arr/.style={-{Stealth[length=2.2mm]}, thick}]
\node[font=\small\bfseries, anchor=west] at (-0.2,4.3) {(a) Test time: write the document from an adapter};
\node[frozen, text width=2.2cm] (ad) at (1.1,2.9) {adapter $\Delta$\\{\scriptsize rank 8, 16 or 64}};
\node[frozen, text width=2.4cm] (base) at (5.0,2.9) {base model $\theta$\\{\scriptsize Qwen3-8B}};
\node[trained, text width=2.0cm] (wr) at (8.6,2.9) {writer $\phi$\\{\scriptsize rank 16}};
\node[textn, text width=2.4cm] (out) at (12.3,2.9) {document $\hat s$};
\node[textn, text width=3.3cm] (txt) at (8.6,1.2) {\EWT{} only: 100 keywords\\+ 30 instruction lines};
\node[fixed, text width=2.6cm] (rd) at (4.0,1.2) {Reader 2\\{\scriptsize reads the weights}};
\draw[arr] (ad) -- node[above,font=\scriptsize]{attach} (base); \draw[arr] (base) -- node[above,font=\scriptsize]{+ writer} (wr);
\draw[arr] (wr) -- node[above,font=\scriptsize]{decode} (out);
\draw[arr, dashed] (ad.south) |- (rd.west); \draw[arr, dashed] (rd) -- (txt); \draw[arr, dashed] (txt) -- (wr);
\node[font=\scriptsize\itshape, text width=3.2cm, align=left] at (12.1,1.2) {prompt: ``Write the SKILL.md document''};
\node[font=\small\bfseries, anchor=west] at (-0.2,-0.5) {(b) Training: only $\phi$ is updated};
\node[textn, text width=2.0cm] (s) at (1.1,-2.0) {document $s$};
\node[frozen, text width=2.2cm] (as) at (4.6,-2.0) {adapter $\Delta(s)$};
\node[frozen, text width=2.7cm] (b2) at (7.8,-2.0) {$\theta\oplus\Delta(s)$};
\node[trained, text width=1.9cm] (w2) at (10.6,-2.0) {writer $\phi$};
\node[textn, text width=2.1cm] (pr) at (12.9,-2.0) {predicted tokens};
\node[box, text width=9.2cm] (loss) at (7.0,-4.0) {next-token loss on the document tokens, $-\log q_{\theta\oplus\Delta(s)\oplus\phi}(s\mid x)$};
\draw[arr] (s) -- node[above,font=\scriptsize]{compile} (as); \draw[arr] (as) -- (b2); \draw[arr] (b2) -- (w2); \draw[arr] (w2) -- (pr);
\draw[arr] (pr.south) |- (loss.east); \draw[arr] (s.south) |- (loss.west);
\draw[arr, dashed] ([xshift=1.2cm]loss.north) -- node[right,font=\scriptsize,pos=0.4]{gradient to $\phi$ only} (w2.south);
\end{tikzpicture}
\caption{The Executing Writer. (a) At test time the adapter is attached to its frozen base model and a small writer $\phi$ makes the adapted
model write the document; \EWT{} also receives the text read by R2 (dashed). (b) Training updates only $\phi$.}
\label{fig:method}
\end{figure}
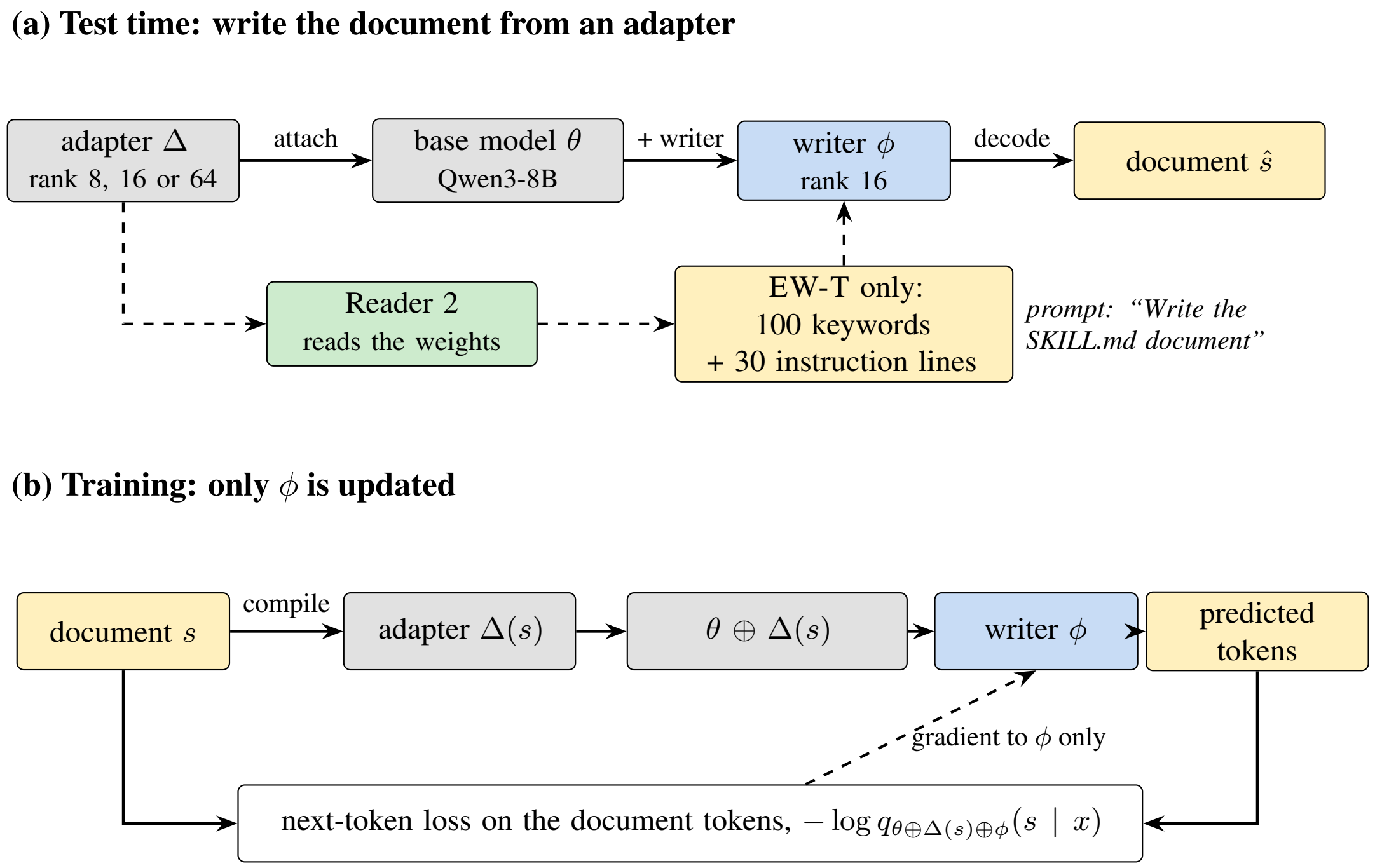

\paragraph{R3: the Executing Writer.} Given $\Delta$ and $\theta$, the writer is a rank-$16$ adapter $\phi$ ($43.6$M parameters) on the same base
model, trained on document--adapter pairs with the base model and $\Delta(s)$ frozen and the loss on the document tokens only (Figure~\ref{fig:method}):
\[
\min_{\phi}\ \mathbb E_{s}\Big[-\log q_{\theta\oplus\Delta(s)\oplus\phi}\big(s\mid x(s)\big)\Big].
\]
The prompt $x(s)$ is the request \emph{``Write the SKILL.md document''}, either alone (\EWA{}, adapter only) or preceded by the keywords and
retrieved lines of R2 with the lines of $s$ itself removed (\EWT{}, adapter and text). \EWT{} is trained on document-fitted adapters and \EWA{}
on task-tuned ones. Documents are decoded greedily with a repetition penalty of $1.1$.

\subsection{Why running the adapter helps: the adapter gap}
\label{sec:info}

An adapted layer updates the hidden state as $h^{(l+1)}=h^{(l)}+F_l(h^{(l)};W_l+\Delta W_l)$, so $\Delta$ affects any output only through the
products $\Delta W_b\,h$. A reader of the numbers must learn a map from $\Delta W$ to text that implicitly simulates what the network does with
$\Delta W$, from a few thousand examples. Running the adapter hands that simulation to the frozen network, which already contains it, and asks
the writer only for a query that makes the adapted model say what it holds. We measure how much each reader takes from the adapter with the
\emph{adapter gap}:
\[
\mathcal G(\mathcal R)=\frac1{|s|}\,\mathbb E_{s,s'}\Big[\log q^{\mathcal R}\big(s\mid x;\Delta(s)\big)-\log q^{\mathcal R}\big(s\mid x;\Delta(s')\big)\Big],
\]
in nats per token, with $s'$ another skill. It is $0$ for a reader that ignores the adapter, and $|s|\,\mathcal G/\ln 2$ is the number of bits the
adapter adds to the document's probability. It is a diagnostic, not a bound on mutual information. For scale, naming one of our $11{,}000$ training
skills takes $\log_2 11{,}000\approx13.4$ bits.

\section{Experimental Setup}
\label{sec:setup}

\begin{table}[!htb]
\centering
\caption{Experiments, models and metrics. Numbers from different rows are comparable only where the setting matches.}
\label{tab:exps}
\footnotesize
\setlength{\tabcolsep}{4pt}
\begin{tabular}{>{\raggedright\arraybackslash}p{3.2cm}>{\raggedright\arraybackslash}p{2.8cm}>{\raggedright\arraybackslash}p{2.7cm}>{\raggedright\arraybackslash}p{2.3cm}>{\raggedright\arraybackslash}p{2.0cm}}
\toprule
Experiment & Base model & Adapters & Skills / tasks & Metric\\
\midrule
Three readers (\S\ref{sec:res-readers}) & Qwen3-8B & document-bearing, rank $8$ & $37$ skills & coverage\\
Transfer and rank (\S\ref{sec:res-rank}) & Qwen3-8B & recitation, rank $16$--$64$ & $37$ + $41$ skills & coverage\\
Behavioural adapters (\S\ref{sec:res-beh}) & Qwen3-8B & behavioural, rank $16$, $64$ & $41$ skills, $323$ tasks & coverage, task score\\
Agent with documents (\S\ref{sec:res-agent}) & Claude Haiku 5.5 agent & recitation, rank $64$ & $4$ benchmark tasks & verifier tests\\
Executing agent (\S\ref{sec:ax}) & Qwen2.5-7B, Qwen3.5-4B & behavioural, rank $16$--$32$ & ALFWorld, $6$ other tasks & task success\\
\bottomrule
\end{tabular}
\end{table}

\paragraph{Models.} All base and actor models are open-weight models: Qwen3-8B and Qwen3-4B~\cite{qwen3}, Qwen2.5-7B-Instruct~\cite{qwen25} and Qwen3.5-4B
(Alibaba), and Llama-3.1-8B-Instruct~\cite{llama3} (Meta). Agents and judges are proprietary models used through their providers: Claude Haiku~5.5
and Claude Sonnet~5.5 (a small and a mid-sized model from Anthropic), run inside Claude Code, Anthropic's
command-line agent harness. The primary coverage judge is JEV-1.13, a judge model that returns a probability for each verdict; Gemini~2.5 Pro and
Gemini~3 Flash (Google) serve as second judges. Open models are served with vLLM~\cite{vllm}. We refer to models by these names throughout; later
versions may behave differently, so we give all prompts and tool definitions in the released code.

\paragraph{Adapters and data.} Document-bearing adapters come from a public skill-to-adapter hypernetwork~\cite{latentskill}: rank $8$ on all seven
projections of all $36$ layers of Qwen3-8B ($252$ blocks, $21.8$M numbers), generated on demand. From its corpus of $171{,}258$ skill documents we
draw $10{,}500$ training documents with strict near-duplicate filtering against the evaluation set and add $500$ documents from an earlier
training set ($11{,}000$ in all), plus $300$ validation and $200$ further held-out documents. The test set has $37$ skills ($9$ from the corpus, $5$ hand-written and $23$ from a split the hypernetwork never saw); seven have near-copies
in the training corpus, so we also report the other $30$.

\paragraph{Baselines.} \emph{Retrieval}: the training document closest to the adapter. \emph{Keyword writer} and \emph{instruction writer}:
Qwen3-4B with a rank-$16$ adapter trained to write the document from the keywords, or from keywords and $30$ retrieved lines. \emph{Soft-prompt
writers}: R1 with a Qwen3-4B or Qwen3-8B writer.

\paragraph{Training and evaluation.} \EWT{} and \EWA{} are trained for one pass over the training set ($1{,}312$ steps at effective batch
$8$, learning rate $2\times10^{-4}$, rank $16$, $\alpha=32$). Differences below about four coverage points are within noise on $37$ skills.

\section{Results on Document-Bearing and Recitation Adapters}
\label{sec:res-doc}

\subsection{Which reader works?}
\label{sec:res-readers}

\begin{table}[!htb]
\centering
\caption{Recovery on the $37$ test skills (percent). Coverage by two judges, coverage with a wrong adapter, coverage on the $30$ skills without
near-copies, and the unsupported share (lower is better). $\dagger$: runs the adapter. \EWT{} reads document-fitted and \EWA{} task-tuned
adapters; the other rows use document-fitted adapters.}
\label{tab:main}
\small
\setlength{\tabcolsep}{3.5pt}
\begin{tabular}{lrrrrr}
\toprule
Method & Judge 1 & Judge 2 & Wrong adapter & 30 skills & Unsupported\\
\midrule
Retrieval of a similar document                    & 26.9 & 35.2 & --- & 15.3 & 66 \\
Keyword writer (weights only)                      & 16.1 & 23.5 & 3.3 & 14.6 & --- \\
Instruction writer (weights only)                  & 22.3 & 33.3 & 1.6 & 20.4 & 54 \\
Soft-prompt writer, 4B (weights only)              & 22.7 & 33.2 & 22.5 & 18.2 & 55 \\
Soft-prompt writer, 8B (weights only)              & 20.9 & 31.3 & 23.1 & 18.2 & 55 \\
\midrule
\EWT$^\dagger$ (adapter and text)                  & \best{43.7} & \best{61.5} & 1.3 & \best{39.9} & \best{39} \\
\EWA$^\dagger$ (adapter only)                      & 38.7 & 59.7 & 1.5 & 37.3 & \best{39} \\
\bottomrule
\end{tabular}
\end{table}

Table~\ref{tab:main} shows that running the adapter nearly doubles recovery. The Executing Writer recovers $38.7\%$ and $43.7\%$ of the key
instructions, against $22.3\%$ for the best weight-only writer and $26.9\%$ for retrieval. This gain comes from the adapter. With another skill's
adapter attached, the executing writers fall to $1.3$--$1.5\%$, although \EWT{} still receives the correct text, so it follows the adapter over the text, as expected for adapters
that store their document. \EWA{} receives no text at all, so everything it writes comes from the adapter. The soft-prompt writers, in contrast, score the same with a wrong adapter, so they do not use it.
The executing writers also invent less: $39\%$ of their units are unsupported by the original, against $54$--$66\%$ for the other methods. On the
$30$ skills without near-copies, retrieval falls to $15.3\%$, while the executing writers keep $37$--$40\%$. Appendix~\ref{app:judge} compares the
two judges: the second is more generous to every method, by $7$ to $21$ points, but under both the executing writers lead.

\subsection{What the weights contain, and why a learned channel stays weak}

The weights do contain the skill's identity and its content words. The retrieval encoder names the correct document for all $37$ test skills
and reaches $100\%$ top-1 on $300$ validation skills, and a single threshold on the keyword score separates an adapter's own document from the
most similar other document with AUC $0.905$ on $537$ unseen adapters. This gives a training-free check that an adapter matches a claimed
skill, though it checks the skill and not its individual instructions.

\begin{table}[!htb]
\centering
\caption{Adapter gap on $100$ validation skills (nats per token): how much the correct adapter lowers the document's loss against a wrong one.}
\label{tab:gap}
\small
\begin{tabular}{lrrrr}
\toprule
Reader & Loss, correct & Loss, wrong & Gap $\mathcal G$ & Bits per 1k tokens \\
\midrule
R1 soft prompt, complete adapter           & 1.220 & 1.221 & 0.001 & $\approx1$ \\
R2 keywords                                & 1.292 & 1.375 & 0.083 & $\approx120$ \\
R2 keywords and instruction lines          & 1.271 & 1.398 & 0.127 & $\approx180$ \\
R3 Executing Writer, \EWT{}                & 0.289 & 3.934 & 3.645 & $\approx5{,}260$ \\
R3 Executing Writer, \EWA{}                & 0.323 & 3.936 & 3.613 & $\approx5{,}210$ \\
\bottomrule
\end{tabular}
\end{table}

Table~\ref{tab:gap} shows where the difficulty lies: in turning the numbers into text. Across five soft-prompt variants the gap never exceeds $0.006$ nats per token,
about one bit for a $1{,}000$-token document, less than the $13.4$ bits needed just to name the skill. The channel can carry information: with eight
training skills the same writer learns to use the adapter completely (loss $0.000$ with the correct adapter, $0.731$ with a wrong one), and on
training skills at scale the gap is still only $0.002$--$0.003$, so the weakness is not overfitting. The keyword and line readers pass on a few
hundred bits and the executing writers several thousand. Measured against no adapter rather than a wrong one, the executing writers'
gap is $1.45$ and $1.33$ nats per token, about $2{,}000$ bits per $1{,}000$ tokens; a wrong adapter lowers the likelihood further because it
points the writer at another document. Against either baseline, running the adapter passes on over ten times more than any weight-only reader. Capacity is not the cause either: a rank-$16$ adapter on
$252$ blocks has $43.6$M parameters, four orders of magnitude more than the roughly $2.5$ bits per token (the writer's loss with no adapter) that a document needs beyond the base model.
Better weight encoders may close part of this gap; with the encoders we tried, running the adapter was far more effective.

\subsection{Transfer to other adapters, and the effect of rank}
\label{sec:res-rank}

To test transfer, we fitted new Qwen3-8B adapters to recite each of the $37$ test skills at ranks $16$, $32$ and $64$ and applied both writers with no
retraining. We did the same for $41$ skills from SkillsBench~\cite{skillsbench}, a public benchmark whose skills and tasks we did not write and which no
reader saw during training. All fitted adapters reproduce their document exactly, so rank does not limit storage.

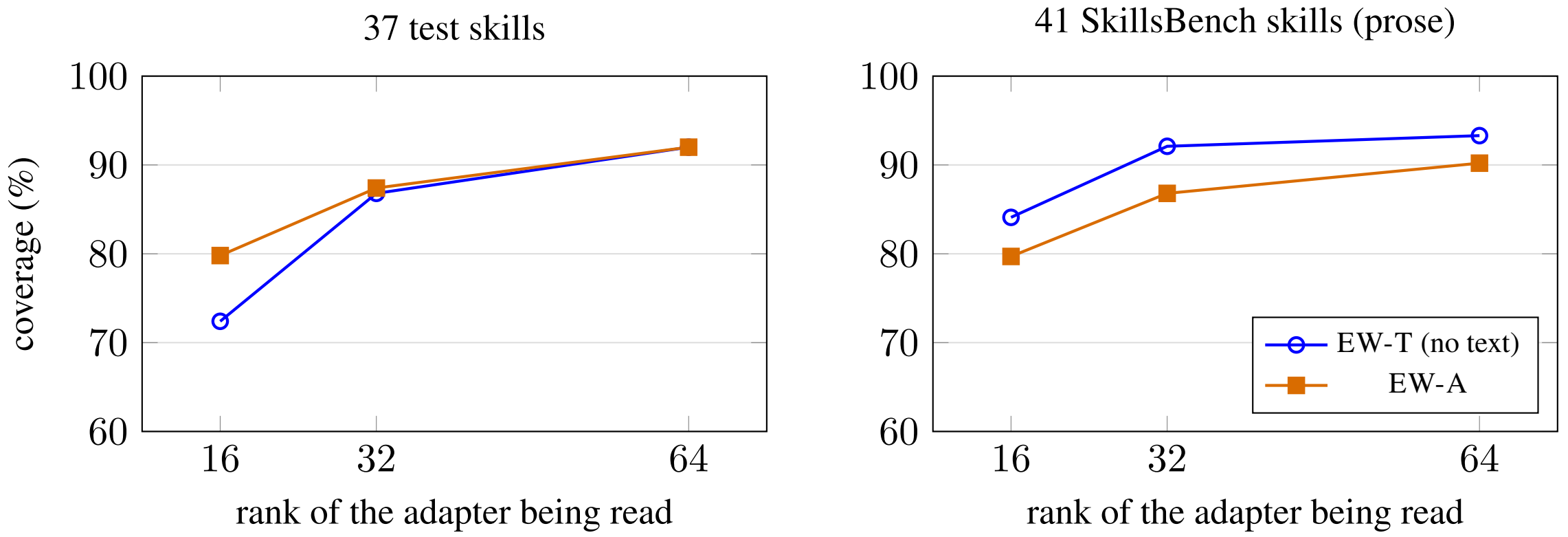
\begin{figure}[!htb]
\centering
\begin{tikzpicture}
\begin{groupplot}[group style={group size=2 by 1, horizontal sep=1.6cm}, width=0.46\linewidth, height=5.0cm,
  xtick={16,32,64}, xmin=8, xmax=72, ymin=60, ymax=100, xlabel={rank of the adapter being read},
  ylabel style={font=\small}, xlabel style={font=\small}, ymajorgrids, grid style={gray!25},
  legend style={font=\scriptsize,at={(0.97,0.05)},anchor=south east}]
\nextgroupplot[title={\small 37 test skills}, ylabel={coverage (\%)}]
\addplot[mark=o,thick,blue] coordinates {(16,72.4) (32,86.8) (64,92.0)};
\addplot[mark=square*,thick,orange!85!black] coordinates {(16,79.8) (32,87.4) (64,92.0)};
\nextgroupplot[title={\small 41 SkillsBench skills (prose)}]
\addplot[mark=o,thick,blue] coordinates {(16,84.1) (32,92.1) (64,93.3)};
\addplot[mark=square*,thick,orange!85!black] coordinates {(16,79.7) (32,86.8) (64,90.2)};
\legend{\EWT{} (no text),\EWA{}}
\end{groupplot}
\end{tikzpicture}
\caption{Recitation adapters are read back more accurately at higher rank. Both writers have rank $16$; only the adapter being read changes. With a wrong
adapter coverage is about $2\%$ (left) and $9$--$13\%$ (right). Full tables in Appendix~\ref{app:rank}.}
\label{fig:rank}
\end{figure}

Figure~\ref{fig:rank} shows that both writers transfer. On the $37$ skills they recover $72$--$80\%$ of the instructions at rank $16$ and $92\%$ at rank $64$; on
the $41$ external skills, $80$--$84\%$ of the prose at rank $16$ and $90$--$93\%$ at rank $64$, with the same numbers for code and structure and for a
judge-free count of identifiers. Most of the gain is in place by rank $32$, close to the effective rank of these updates (about $40$ directions by
participation ratio). These adapters were fitted to recite their document, so the experiment shows transfer, not that behaviour alone can be read.

\subsection{Behavioural adapters}
\label{sec:res-beh}

To test the case that matters in practice, we fitted adapters on the $41$ external skills \emph{without} the document in any training text. For
each skill the base model writes about $240$ requests the skill would help with and answers them with the skill in its prompt; a LoRA is trained to
give the same answers from the request alone. The fitted adapters reach an answer loss close to the prompted model's. We applied both writers to
them unchanged.

\begin{table}[!htb]
\centering
\caption{Document fidelity on the $41$ external skills (percent). ``Same'' is a second judge comparing whole documents.}
\label{tab:beh-doc}
\small
\setlength{\tabcolsep}{4pt}
\begin{tabular}{llrrrrrr}
\toprule
 & & \multicolumn{3}{c}{Coverage of original} & \multicolumn{2}{c}{Precision} & Whole doc.\\
\cmidrule(lr){3-5}\cmidrule(lr){6-7}\cmidrule(lr){8-8}
Method & Rank & Prose & Struct. & Ident. & Prose & Struct. & Same\\
\midrule
\multicolumn{8}{l}{\emph{Recitation adapters}}\\
\EWT{} & 16 & 84.1 & 85.1 & 85.1 & 85.8 & 65.5 & 83\\
\EWT{} & 64 & \best{93.3} & \best{94.0} & \best{94.1} & \best{94.9} & \best{72.0} & \best{90}\\
\EWA{} & 16 & 79.7 & 80.1 & 78.0 & 84.5 & 63.2 & 73\\
\EWA{} & 64 & 90.2 & 90.2 & 91.2 & 94.2 & 70.0 & 86\\
\midrule
\multicolumn{8}{l}{\emph{Behavioural adapters}}\\
\EWT{} & 16 & 0.9 & 2.4 & -- & 19.3 & 15.3 & 0\\
\EWT{} & 64 & 1.1 & 2.7 & -- & 19.0 & 14.1 & 1\\
\EWA{} & 16 & 1.3 & 4.3 & -- & 20.9 & 11.0 & 1\\
\EWA{} & 64 & 1.9 & 4.8 & -- & 20.1 & 13.5 & 2\\
\midrule
Wrong adapter (both writers) & 64 & 8.3 & 12.2 & -- & 13.7 & 11.0 & 5\\
\bottomrule
\end{tabular}
\end{table}

Table~\ref{tab:beh-doc} shows that the trained writers do not read behavioural adapters. From them both writers recover $1$--$2\%$ of the prose
and $2$--$5\%$ of the structure, below the wrong-adapter control, and the second judge finds $0$--$2\%$ of the instructions present. These writers
were trained only on adapters that recite their document. Appendix~\ref{app:ewb} reports a writer trained on behavioural adapters instead: it
depends on the adapter (a gap of $0.7$ nats per token) but still recovers only about $7\%$ with $45$ training adapters, which it memorises.

To see what the documents and adapters \emph{do}, we wrote $323$ tasks ($8$ per skill) that each require $3$ to $6$ specific behaviours taken from
the original, and scored the base model's answers under each condition with the primary judge. Table~\ref{tab:beh-task} reports the share of the
original's gain over no skill that each condition restores.

\begin{table}[!htb]
\centering
\caption{What documents and adapters do on $323$ skill-dependent tasks (base-model score, percent). Gain recovered is relative to the original.}
\label{tab:beh-task}
\small
\setlength{\tabcolsep}{5pt}
\begin{tabular}{llrrr}
\toprule
Condition & Rank & All tasks & Hardest $91$ tasks & Gain recovered (\%)\\
\midrule
No skill & -- & 55.5 & 23.2 & 0\\
Original skill in prompt & -- & 77.0 & 65.9 & 100\\
\midrule
\multicolumn{5}{l}{\emph{Recovered document in prompt, from recitation adapters}}\\
\EWT{} & 16 & 73.6 & 59.2 & 84\\
\EWT{} & 64 & 75.1 & 60.8 & \best{91}\\
\EWA{} & 16 & 71.5 & 56.0 & 74\\
\EWA{} & 64 & 73.8 & 59.1 & 85\\
\midrule
\multicolumn{5}{l}{\emph{Recovered document in prompt, from behavioural adapters}}\\
\EWT{} & 64 & 55.7 & 30.1 & 1\\
\EWA{} & 64 & 54.6 & 28.7 & $-4$\\
\midrule
\multicolumn{5}{l}{\emph{Behavioural adapter attached, no document}}\\
Adapter itself & 16 & 68.0 & 50.8 & 58\\
Adapter itself & 64 & 65.6 & 50.0 & 47\\
\bottomrule
\end{tabular}
\end{table}

Documents recovered from recitation adapters restore $74\%$ to $91\%$ of the original's gain on all tasks, and $85$--$91\%$ at rank $64$; the
standard error of each paired difference to the original is $0.7$--$1.3$ points. The hardest $91$ tasks, those with a no-skill score of at most
$0.4$, show the same pattern. Behavioural adapters behave differently. Attached with no document, they restore $47$--$58\%$ of the gain, so they
carry the skill; but the documents the writers produce from them restore none of it ($-4$ to $+1\%$). The rank-$16$ adapters restore somewhat more than
rank $64$ ($58\%$ against $47\%$); we did not study why.

\subsection{An agent using recovered documents}
\label{sec:res-agent}

Tasks we wrote ourselves might favour our documents, so we also tested them on SkillsBench tasks, whose verifiers the benchmark authors wrote.
The agent is Claude Haiku~5.5 in the benchmark's own Docker sandbox. Of $17$ pilot tasks we kept the four that Haiku fails without a skill and
that the reference solution passes, and ran each $8$ times per arm with the full skill text in the prompt (Table~\ref{tab:agent}). On the 3D-model
parser task, every skill takes the agent from $0/8$ to $8/8$. On intrusion detection, the original and the \EWT{} reconstruction both reach $5/8$,
while \EWA{} gets $0/8$. Over all runs, the share of verifier tests passed rises from $59.7\%$ with no skill to $85.0\%$ with the original and to
$82.5$--$83.1\%$ with the reconstructions; the differences from the original, about $-2$ points, have intervals of about $-12$ to $+7$. Because the
agent is a different model from the adapter's base, these runs also show a recovered document working across models. Four tasks, two of which
respond to a skill, make a small test. Appendix~\ref{app:examples} adds $41$ tasks we wrote: where no skill was needed the reconstructions never
did worse than the original, and on one of the three skill-dependent tasks \EWT{} fully restored the original's effect.

\begin{table}[!htb]
\centering
\caption{Claude Haiku 5.5 as an agent on four SkillsBench tasks: runs passing all verifier tests (of $8$) and the mean share of tests passed.}
\label{tab:agent}
\small
\setlength{\tabcolsep}{5pt}
\begin{tabular}{lrrrrrr}
\toprule
Skill in prompt & Intrusion det. & 3D parser & Finance QA & PDDL plan & All runs & Tests (\%)\\
\midrule
None & 0/8 & 0/8 & 0/8 & 3/8 & 3/32 & 59.7\\
Original & 5/8 & 8/8 & 2/8 & 2/8 & 17/32 & \best{85.0}\\
\EWT{} reconstruction & 5/8 & 8/8 & 1/8 & 1/8 & 15/32 & 82.5\\
\EWA{} reconstruction & 0/8 & 8/8 & 1/8 & 4/8 & 13/32 & 83.1\\
\bottomrule
\end{tabular}
\end{table}

\begin{findings}
\begin{itemize}
\item Running the adapter is what makes the difference: the Executing Writer recovers $39$--$44\%$ of a held-out skill's key instructions, about twice the best weight-only writer, and falls to $1.3$--$1.5\%$ with a wrong adapter.
\item The weights reveal which skill an adapter holds and its content words; a learned map from weights to text passed on about one bit.
\item Recovered documents work: $85$--$91\%$ of the original's task gain at rank $64$, and $82$--$83\%$ of verifier tests for an agent against $85\%$
with the original.
\item Behavioural adapters remain much harder for trained writers ($1$--$2\%$ of instructions, about $7\%$ for a writer trained on them), although the adapters themselves carry the skill.
\end{itemize}
\end{findings}

\section{Reading Behavioural Adapters with an Executing Agent}
\label{sec:ax}

A behavioural adapter never saw its document, so there is no stored text to extract; what it holds is a change in behaviour. This section tests a
reader that needs no training: an agent that runs the base and adapted models on real task instances, looks at where they act differently, and
writes a skill from those differences. We ask three questions, all judged by tasks solved. Does the skill recover the adapter's benefit? Does it
come from the adapter, or would the agent have written it anyway? Does it work on models the adapter cannot be attached to? Every experiment in this
section, including its pass criteria, was fixed in a written plan before it ran.

\subsection{Setup}
\label{sec:ax-setup}

\paragraph{The agent.} The writer is a Claude model running headless in Claude Code, in a sandbox where its only command is a tool, \texttt{fx},
that wraps the models. With it the agent can play a development game with the base or the adapted model, play both on the same game and see the
first step where they differ, ask both models for their next action at a chosen state, try a draft skill on the base model, and score a draft.
All of these tools observe the models' outputs; two further tools that expose internal signals (Appendix~\ref{app:alf}) were available but
never used, so this reader works from behaviour. Test instances never pass through the tool, and a canary file placed beside the test results never appeared in any transcript. Skills are limited
to about $800$ words. We call one independent run of the agent, from an empty sandbox to a finished skill, a \emph{session}. Sessions of the same arm
can differ a lot, so we run several per arm and report intervals that resample both test items and sessions.

\paragraph{Scoring signals.} In the main one-session arm the agent never sees a reward: its score is how often the base model with its draft picks the
same action as the adapted model at about $500$ development states, so anything it learns about the task it learns from the adapter. In the
\emph{reward loop}, the writer instead sees task success on $60$ development games, revises its drafts over up to five rounds, and the final round
is chosen on $80$ further development games that the writer never sees; the test games are played once per final skill. A linter rejects drafts that name specific objects (``drawer 2''), copy long passages or exceed $800$ words. An audit of the test wins,
described in Appendix~\ref{app:loop}, found no exploits.

\paragraph{Controls.} The agent is also run with a \emph{wrong adapter} (from another task), with the \emph{task hidden}, with only the \emph{task text},
and with ten expert \emph{demonstrations} instead of the adapter. If a control scores like the real arm, the skill did not come from the adapter.

\paragraph{Tasks.} The main benchmark is ALFWorld~\cite{alfworld}, a text-based household game in which an agent must, for example, find a mug, heat it
and put it on a shelf: $134$ unseen test games and separate development games. The actor is Qwen2.5-7B-Instruct, playing greedily. Its LoRA (rank
$16$, one epoch on $1{,}477$ expert games) wins $81.3\%$ of test games against $25$--$26\%$ for the base model. We also use a docker-compose
adapter that we had trained earlier for a separate project, with no thought of reading it back (Appendix~\ref{app:dc}), five synthetic and tool-calling tasks on Qwen3.5-4B (Appendix~\ref{app:fresh}), and ALFWorld LoRAs
trained to follow hidden rules (Section~\ref{sec:ax-dep}). We report the share of the LoRA's gain a skill recovers,
$R=(\text{skill}-\text{base})/(\text{LoRA}-\text{base})$.

\subsection{How much of the LoRA's gain does a skill recover?}
\label{sec:ax-alf}

\begin{table}[!htb]
\centering
\caption{ALFWorld, $134$ unseen games, Qwen2.5-7B-Instruct as actor. Each skill is given to the base model. $R$: share of the LoRA's gain recovered.}
\label{tab:axalf}
\small
\setlength{\tabcolsep}{4pt}
\begin{tabular}{llrrrr}
\toprule
 & Arm & Sessions & Success & Range & $R$\\
\midrule
\multicolumn{6}{l}{\emph{References}}\\
 & LoRA & -- & \best{0.813} & & 1.00\\
 & Base model (one session / loop protocol)$^a$ & -- & 0.254 / 0.261 & & 0.00\\
 & Three in-context expert games$^a$ & -- & 0.709 / 0.679 & & 0.81 / 0.76\\
\midrule
\multicolumn{6}{l}{\emph{One session, Haiku writer, no reward}}\\
 & Agent reads the adapter & 6 & 0.581 & 0.42--0.72 & 0.58\\
 & \quad with dev reward instead of agreement & 3 & 0.592 & 0.55--0.62 & 0.60\\
 & Agent reads 10 demonstrations & 6 & 0.588 & 0.46--0.74 & 0.60\\
 & Agent with a wrong adapter & 6 & 0.521 & 0.30--0.64 & 0.48\\
 & Agent with the task hidden & 6 & 0.496 & 0.02--0.66 & 0.43\\
 & Agent with the task text only & 6 & 0.113 & 0.07--0.15 & $-0.25$\\
\midrule
\multicolumn{6}{l}{\emph{Reward loop, Sonnet writer, five rounds}}\\
 & Loop with the adapter & 12 & 0.653 & 0.54--0.69 & 0.71\\
 & Loop without any adapter & 12 & \best{0.673} & 0.59--0.81 & \best{0.75}\\
 & Loop with 10 demonstrations & 3 & 0.649 & 0.58--0.72 & 0.70\\
 & Loop with a wrong adapter & 2 & 0.660 & 0.60--0.72 & 0.72\\
 & Loop with the adapter, Haiku writer & 4 & 0.618 & 0.56--0.69 & 0.65\\
 & One round with the adapter, agreement score & 4 & 0.483 & 0.27--0.58 & 0.40\\
\midrule
 & Loop skill $+$ three in-context games (adapter / none) & 12 / 12 & 0.795 / 0.778 & & 0.97 / 0.94\\
\bottomrule
\end{tabular}

\smallskip
\begin{minipage}{0.95\linewidth}\footnotesize
$^a$The two groups were run separately with step caps of $50$ and $52$; base model and in-context references are given for each. The LoRA scores
$0.813$ under both. ``Range'' is the lowest and highest session.
\end{minipage}
\end{table}

Table~\ref{tab:axalf} shows that running the adapter gives a good skill and that a reward loop gives a better one. In one session with no reward, the agent lifts
the base model from $0.254$ to $0.581$, $58\%$ of the LoRA's gain. Its skills move the base model's actions measurably closer to the LoRA's (agreement
at matched states $0.646$, against $0.581$ for the base model), and watching the models matters: an agent that sees only the task text and the reward
writes skills that make the actor worse ($0.113$). With a reward loop, a stronger writer and five rounds, the skill alone reaches $0.653$--$0.673$
($R=0.71$--$0.75$). Selection on unseen development games carried over to the test games, which score only about $3$ points lower.
The pre-set goal of $R\ge0.80$ for a skill alone was not reached, and no skill reached the LoRA. Combining a loop skill with three in-context
games reaches $0.78$--$0.80$ ($R\approx0.95$), against $0.68$ for the games alone; this is a different protocol, since it adds demonstrations to the prompt.

The controls come close, however. In one session, a wrong adapter ($0.521$), a hidden task ($0.496$; $0.59$ without one session whose skill made the
actor repeat \texttt{look}) and ten demonstrations ($0.588$) all score near the real arm. In the reward loop, the same loop without any adapter does
as well as with it: the difference is $-2.0$ points, with interval $[-7.6,+3.8]$ (the pre-set goal was $+5$). This LoRA imitates an expert, so it holds
what any competent household agent does, and a capable writer can infer most of that from the task and the reward. Session spread is large: one
loop session without the adapter reached $0.806$, close to the LoRA, but sessions of the same arm span $22$ points.

\subsection{Does the skill come from the adapter?}
\label{sec:ax-dep}

\begin{table}[!htb]
\centering
\caption{Skills written from the real adapter versus a control, across tasks. Intervals resample items and sessions; \checkmark: excludes $0$.}
\label{tab:axdep}
\small
\setlength{\tabcolsep}{4pt}
\begin{tabular}{>{\raggedright\arraybackslash}p{3.6cm}>{\raggedright\arraybackslash}p{3.3cm}rrl}
\toprule
Task (control) & What the adapter learned & Real & Control & Difference [$95\%$ CI]\\
\midrule
Docker run$\rightarrow$compose (wrong) & arbitrary output conventions & 0.212 & 0.000 & $+0.212$ [$+0.127$, $+0.315$] \checkmark\\
S4 ticket triage (wrong) & arbitrary rule & 0.873 & 0.645 & $+0.228$ [$+0.148$, $+0.323$] \checkmark\\
S2 fee calculation (wrong) & arithmetic rule & 0.467 & 0.394 & $+0.072$ [$-0.034$, $+0.176$]\\
S1 house format (wrong) & learnable from inputs & 0.908 & 0.893 & $+0.015$ [$-0.028$, $+0.063$]\\
ALFWorld, one session (wrong) & expert household play & 0.581 & 0.521 & $+0.060$ [$-0.061$, $+0.184$]\\
ALFWorld, loop (no adapter) & expert household play & 0.653 & 0.673 & $-0.020$ [$-0.076$, $+0.038$]\\
\midrule
ALFWorld hidden rules$^b$ (decoy) & one-step habits & 0.029 & 0.000 & $+0.029$ [$+0.006$, $+0.061$] \checkmark\\
\bottomrule
\end{tabular}

\smallskip
\begin{minipage}{0.95\linewidth}\footnotesize
$^b$Rule success, pooled over two rules: the share of games won while following the rule.
\end{minipage}
\end{table}

Table~\ref{tab:axdep} shows that the adapter is the source of the skill when its behaviour cannot be guessed. On docker, skills from the real adapter reach
$0.212$ exact match against $0.000$ in every wrong-adapter session: the reference output follows conventions, such as a fixed project name and a null
default network, that the base model and the agent have no way to know. On the synthetic triage task S4 the real adapter gives $0.873$ against
$0.645$. Where the behaviour is guessable or learnable from the inputs (ALFWorld, S1, S2) the intervals include zero.

\paragraph{Hidden rules.} To test the unguessable case on ALFWorld itself, we trained two more LoRAs on the same expert games with one arbitrary step
inserted: LoRA-A says \texttt{inventory} just before the winning action, and LoRA-B says \texttt{look} right after the first successful \texttt{take}.
They follow their rules in $87\%$ and $85\%$ of test games, while the base model never does. Each skill is scored under both rules, so a skill
written from LoRA-A is the real arm for rule A and a competent decoy for rule B. The agents found the rules: the skill text stated the right rule in
$10$ of $12$ sessions and the wrong rule in none, while agents reading demonstrations or the base model never stated either. But the actor rarely
carried them out: rule success was only $0.5\%$ for rule A and $5.3\%$ for rule B. Even a skill that states the rule outright makes the 7B actor follow it in only $3\%$ (A) and
$19\%$ (B) of games, so this test is limited by the actor more than by the reader.

\paragraph{Rules the actor can follow.} We then searched for rules the actor does follow from text and found
two ($65\%$ and $53\%$ compliance when stated): \texttt{inventory}, or examining the object, right after the first take. With LoRAs for these rules and
the reward loop, skills reached $9$--$41\%$ rule success for the first rule and $0$--$22\%$ for the second, against $78\%$ for the LoRAs and $0\%$ for the decoy rule in every session (details in Appendix~\ref{app:rules}). The agent can read an arbitrary rule from a LoRA, but turning it into a skill that a small
actor follows reliably remains hard.

\subsection{Portability and what the skills add}
\label{sec:ax-port}

\begin{table}[!htb]
\centering
\caption{The same skill documents, unchanged, on other actors (ALFWorld, $134$ unseen games). The LoRA cannot be attached to any of them.}
\label{tab:axport}
\footnotesize
\setlength{\tabcolsep}{4pt}
\begin{tabular}{lrrrrr}
\toprule
Actor & Base & One-session skill & Loop, adapter & Loop, no adapter & Three games\\
\midrule
Qwen2.5-7B (the LoRA's base) & 0.254 & 0.612 & 0.676 & 0.658 & 0.709\\
Llama-3.1-8B-Instruct & 0.052 & \best{0.545} & 0.454 & 0.405 & 0.269\\
Qwen3.5-4B & 0.022 & \best{0.664} & 0.385 & 0.564 & 0.396\\
Claude Haiku 5.5 (18 games) & 0.944 & 1.000 & n.r. & n.r. & n.r.\\
\bottomrule
\end{tabular}

\smallskip
\begin{minipage}{0.95\linewidth}\footnotesize
One-session skill: the session chosen on development games. Loop columns: mean of the first five sessions of each arm, with no selection.
n.r.: not run.
\end{minipage}
\end{table}

Table~\ref{tab:axport} shows the main advantage of a document: it moves to any model. The one-session skill lifts Llama-3.1-8B from $0.052$ to $0.545$ and
Qwen3.5-4B from $0.022$ to $0.664$, gains of $49$ and $64$ points, larger than its $36$ on the home model; three in-context games transfer much less
($+22$ and $+37$). Part of this is formatting: both models fail mostly through invalid commands, which any skill repairs. Skills from the reward loop also transfer, though less than the one-session skill (Llama-3.1-8B $0.41$--$0.45$, Qwen3.5-4B $0.39$--$0.56$,
where two loop sessions with the adapter failed), perhaps because the loop fits them to the home actor. Loop skills written without any adapter
transfer as well as those written with one, so what moves between models is the document itself. Claude Haiku is
already at ceiling, so its row says little.

\begin{table}[!htb]
\centering
\caption{Success by ALFWorld task type on the home actor (reward-loop skills, mean of $12$ sessions). Games per type in parentheses.}
\label{tab:axtype}
\small
\setlength{\tabcolsep}{4pt}
\begin{tabular}{lrrrrrr}
\toprule
 & Pick (24) & Examine (18) & Clean (31) & Heat (23) & Cool (21) & Pick two (17)\\
\midrule
Base model & 0.50 & 0.06 & 0.16 & 0.39 & 0.19 & 0.24\\
LoRA & \best{1.00} & 0.56 & \best{1.00} & \best{1.00} & \best{1.00} & 0.00\\
Loop skill, adapter & 0.84 & \best{0.64} & 0.69 & 0.75 & 0.54 & 0.35\\
Loop skill, no adapter & 0.83 & 0.61 & 0.70 & 0.73 & 0.65 & \best{0.43}\\
\bottomrule
\end{tabular}
\end{table}

Table~\ref{tab:axtype} shows that the skills and the LoRA succeed on different games. The LoRA solves every pick, clean, heat and cool game and no game that
asks for two objects; the skills solve $35$--$43\%$ of those and more of the examine-in-light games than the LoRA, while losing a quarter to a half of
the clean, heat and cool games. A written rule such as ``after placing the first object, go back for the second'' is easy to state and easy for the actor
to follow; a long fixed sequence of steps is where the LoRA's training pays off.

\paragraph{Text on top of the LoRA.} Adding a skill to the LoRA helps only if it is short. A $148$-word skill raises the LoRA from $0.813$ to $0.858$
($+4.5$ points, interval over games $[+1.5,+8.2]$), and three in-context games give $0.836$; the full $847$-word skill lowers it to $0.716$, and the
reward-loop skills to $0.74$--$0.77$. The LoRA was trained with no text in its prompt, and a long text moves it away from what it saw in training;
Appendix~\ref{app:alf} gives every placement we tried.

\subsection{What the weights show}
\label{sec:ax-weights}

The agent above runs the models. We also asked whether the hidden rules can be seen inside the model. A logit lens on the singular directions
of the weight update is not specific: the rule words rank first for the plain ALFWorld LoRA too, as do most command words. A lens on the
\emph{activations} is specific. We take the difference between the adapted and base model's hidden states at $200$ development states, subtract the
mean over the three ALFWorld LoRAs, and read it through the output embedding: \texttt{inventory} ranks first at layer $26$ for LoRA-A ($447$ and
$5{,}113$ for the other two), and \texttt{look} first at layer $24$ for LoRA-B ($222$ and $4{,}400$); the layer was chosen after the fact. Ablations
place the rule in the MLP weights of the middle and upper layers: keeping only the MLP updates of LoRA-A preserves $92\%$ rule success on development
games, attention alone $7\%$, and the first nine layers alone none. These lenses show that a hidden rule leaves a specific trace inside the adapted model, and where that trace sits; turning such traces into a
complete, usable skill without running the model is left for future work (Appendix~\ref{app:weights} gives the details).

\begin{findings}
\begin{itemize}
\item An agent that runs the base and adapted models writes skills that recover $58\%$ of an ALFWorld LoRA's gain in one session and $71\%$ with a reward
loop, and its skills transfer to models the LoRA cannot be attached to (Llama-3.1-8B $0.05\to0.55$, Qwen3.5-4B $0.02\to0.66$).
\item The adapter is the source of the skill when its behaviour cannot be guessed (docker $0.21$ against $0.00$; an arbitrary rule $0.87$ against $0.65$).
On ALFWorld a reward loop with no adapter does as well ($-2.0$ points, $[-7.6,+3.8]$).
\item Hidden one-step rules are read by the agent ($10$ of $12$ sessions) but rarely followed by a 7B actor from text; rules the actor can follow are
partly transferred ($9$--$41\%$ against $0\%$ for the decoy).
\item Skills and the LoRA fail on different task types; short text helps the LoRA and long text hurts it. A centred activation lens finds the hidden
rule's word; a lens on the weight update alone does not.
\end{itemize}
\end{findings}

\section{Discussion}
\label{sec:discussion}

\paragraph{What the results support.} An adapter's document can be recovered to a useful degree from the adapter and its base model, and the
most effective route we found is to let the adapted model do the reading. The wrong-adapter control, the retrieval baseline and the
unsupported-content measure show that the recovered text depends on the adapter and is more faithful than the alternatives. Recovery is far from
exact on the hypernetwork's own adapters ($44\%$ of key instructions, $39\%$ of units unsupported) and high on adapters that recite their document
($90$--$93\%$ of prose at rank $64$), where higher rank helped most.

\paragraph{Why recitation and behaviour differ.} A recitation adapter holds the document's tokens, so the adapted model can be asked to produce
them. A behavioural adapter holds the effect of the document, and nothing in its training asked it to write the document down. Our trained writers
saw only the first kind, so their weak result on the second is a statement about their training data as much as about the method; in Appendix~\ref{app:ewb}, a writer trained on a small bank of behavioural adapters improved slightly and then overfitted. The executing agent sidesteps training and
reads behaviour directly, but on a task like ALFWorld a capable writer can produce the same skill without the adapter. The adapter's information is
most valuable exactly where behaviour is arbitrary.

\paragraph{Why recovery is worth having.} Even where a LoRA scores higher than any document, the document keeps properties the adapter lacks: it can be
read before use, edited without retraining, merged with other skills and given to any model. A skill an agent wrote from a Qwen2.5 LoRA lifted Llama-3.1-8B
from $5\%$ to $55\%$, and a skill written for a 4B model's fee-calculation task raised Claude Haiku from $0\%$ to $96\%$ (Appendix~\ref{app:fresh}), a model on which the LoRA cannot be loaded at all. Recovery is also a way to move a skill off the model it was compiled for.

\paragraph{Limitations.} Most positive results are on adapters that store their document; the hypernetwork's adapters are one family and Qwen3 the
only base family for the trained writers. Coverage relies on LLM judges with no human labels: the two judges agree on $272$ of $485$ verdicts and the
second is $7$ to $21$ points more generous, so absolute numbers are less certain than the order of methods. The main skill set has $37$ skills and the
benchmark agent test four tasks. The executing-agent results rest on one main benchmark and one home actor, with $2$ to $12$ sessions per arm, and
the spread between sessions is often larger than the gap between arms. The 7B actor limits what any skill can show for hidden rules. Finally, the original
document is an imperfect target: an adapter may not have learned every instruction, and different wording can have the same effect.

\paragraph{Future work.} The main open problem is behaviour that is both arbitrary and followable: an adapter whose behaviour cannot be
guessed is where reading it matters most, yet a small actor often fails to carry out the rule once it is written down. Three directions follow. First, breadth: more benchmarks (SkillsBench, WebShop, ScienceWorld, $\tau$-bench), more
adapters trained by others, and a second base-model family, with human labels for a sample of documents. Second, the unguessable case: adapters whose
behaviour is arbitrary but easy to follow (a format, a required field, a naming convention), the reward loop run on such adapters to see whether
the adapter then adds what it did not add on ALFWorld, and stronger actors. Third, better readers: writers trained on several hundred behavioural
adapters with early stopping, agents that combine execution with activation lenses on the middle layers, and searches that use the adapter's own
answers to score candidate documents.

\paragraph{Broader impact.} Recovering a document from an adapter helps audit and edit adapters before they are run, and lets a user check that an adapter
matches its stated skill. The same capability lets someone who holds a released adapter reconstruct the skill it compiles, which may be proprietary,
so releasing only weights should not be treated as keeping a skill private. Recovered documents are partial and contain unsupported content, and
should be checked before they are trusted.

\section{Conclusion}

Skills compiled into adapters are cheap to run but lose what makes a skill useful: a document that can be read, edited, audited and given to
another model. We asked whether that document can be written back, and judged the answer by the instructions it recovers and the tasks it
solves. Reading the weights alone recovered a skill's identity and its words, but little of its instructions. Running the adapter worked. A
small writer trained on top of the adapted model recovers $39$--$44\%$ of a held-out skill's key instructions, about twice the best weight-only writer, and almost nothing with a wrong adapter. On adapters that store their document it recovers $90$--$93\%$ of the prose, and
the recovered documents restore $85$--$91\%$ of the original's task gain and help an agent on benchmark tasks nearly as much as the original.
For adapters trained on behaviour alone, an agent that runs the base and adapted models writes skills that recover $58$--$71\%$ of an ALFWorld
LoRA's gain, read rules that no one could guess, and carry the skill to models the LoRA cannot be loaded on, lifting Llama-3.1-8B from $5\%$
to $55\%$. Together these results show that an adapter need not be a sealed artefact: run inside its base model, it can be turned back into a
document that people can read and other models can use.


\appendix
\section{Evaluation details}
\label{app:judge}

\paragraph{Judges.} The primary judge is a judge model (JEV-1.13) that returns a probability for each verdict on each instruction, so coverage is an
expectation. The second judge (Gemini~2.5 Pro) returns one verdict per instruction. The two agree on $272$ of $485$ instruction verdicts. Most
disagreements are a ``full'' from the second judge against a ``partial'' from the primary; in a random audit of $12$ disagreements the second judge was
too generous in about $10$ (for example, accepting ``cherry-pick supporting data'' as stating ``avoid cherry-picking''), so its coverage runs about $20$
points higher. Both judges were reached through a commercial API router.

\paragraph{Frozen requirements.} The key instructions of each original are extracted once, before any method is scored, and reused for every method
and judge. For the external skills no model extracts requirements: the prose list is every instruction sentence of the original ($6$--$60$ words, with
code, tables and headings removed) and the structure list is every code block, table and heading. Requirements are judged in chunks of $15$ with the
whole original and the whole candidate in each call ($817$ prose and $931$ structure units over $41$ skills). Units are not weighted by importance and
may repeat each other, so these numbers compare methods on identical questions rather than giving an absolute share of the skill.

\paragraph{Whole-document judge.} Table~\ref{tab:beh-doc} also uses a judge that needs no checklist: Gemini~3 Flash (temperature $0$) reads the complete
original and reconstruction, lists every instruction, command, parameter or warning of the original, marks each as same, partial, missing or
contradicted, and states whether an agent given the reconstruction would behave like one given the original ($76$--$90\%$ of skills for recitation
adapters at rank $64$).

\paragraph{Unsupported content.} Units of the candidate are split by rule, not extracted by a model, and the judge is asked whether the original
supports each sampled unit.

\paragraph{Intervals.} For the executing agent, intervals are a paired bootstrap that resamples test items and, within each arm, sessions. Intervals
over items alone are narrower and are not used for claims. All plotted values are drawn from the tables; Figures~\ref{fig:pipeline}
and~\ref{fig:overview}--\ref{fig:method} are schematic.

\section{Transfer and Rank: Full Results}
\label{app:rank}

\begin{table}[h]
\centering
\caption{Executing Writers on adapters fitted to recite each of the $37$ test skills (coverage, percent; whole-document match probability).}
\label{tab:rank}
\small
\setlength{\tabcolsep}{4pt}
\begin{tabular}{lrrrrrrrrr}
\toprule
 & \multicolumn{3}{c}{Coverage} & \multicolumn{3}{c}{Wrong adapter} & \multicolumn{3}{c}{Whole-doc.\ match}\\
\cmidrule(lr){2-4}\cmidrule(lr){5-7}\cmidrule(lr){8-10}
Writer & $r=16$ & $r=32$ & $r=64$ & $r=16$ & $r=32$ & $r=64$ & $r=16$ & $r=32$ & $r=64$\\
\midrule
\EWT{} (no text) & 72.4 & 86.8 & \best{92.0} & 2.3 & 1.9 & 1.7 & 60.7 & 79.3 & 82.7\\
\EWA{}           & 79.8 & 87.4 & \best{92.0} & 1.9 & 1.9 & 2.0 & 52.2 & 69.3 & 76.1\\
\bottomrule
\end{tabular}
\end{table}

\begin{table}[h]
\centering
\caption{Executing Writers on $41$ SkillsBench skills with recitation adapters (percent). Identifier recall is counted by rule, with no judge.}
\label{tab:sb}
\small
\setlength{\tabcolsep}{5pt}
\begin{tabular}{llrrrr}
\toprule
Writer & Rank & Prose cov. & Structure cov. & Identifier recall & Match prob.\\
\midrule
\EWT{} (no text) & 16 & 84.1 & 85.1 & 85.1 & 74.3\\
\EWT{} (no text) & 32 & 92.1 & 93.4 & \best{94.4} & 79.1\\
\EWT{} (no text) & 64 & \best{93.3} & \best{94.0} & 94.1 & \best{81.2}\\
\EWA{} & 16 & 79.7 & 80.1 & 78.0 & 52.1\\
\EWA{} & 32 & 86.8 & 86.9 & 87.7 & 64.8\\
\EWA{} & 64 & 90.2 & 90.2 & 91.2 & 73.1\\
\bottomrule
\end{tabular}
\end{table}

The fitted adapters reach $100\%$ token accuracy at every rank ($r=16$ already stores a $1{,}451$-token document exactly after $25$ steps). Coverage rises
with rank and levels off near the effective rank of the updates (participation ratio $11.6$ at $r=16$ and $40.7$ at $r=64$). The wrong-adapter control
is higher on the external skills ($9$--$13\%$) because these code-heavy documents share boilerplate. The three checklists agree within about one point,
so leaving code out of the prose list did not inflate the scores.

\section{Examples, Self-Authored Agent Tasks and Cost}
\label{app:examples}

\begin{figure}[h]
\centering
{\fontsize{6}{7}\selectfont
\begin{minipage}[t]{0.485\linewidth}
\textbf{Original} (Zustand store skill)
\begin{verbatim}
## Basic Store
```typescript
import { create } from "zustand";
interface CounterStore {
  count: number;
  increment: () => void;
  decrement: () => void;
  reset: () => void;
}
const useCounterStore = create<CounterStore>((set...
\end{verbatim}
\end{minipage}\hfill
\begin{minipage}[t]{0.485\linewidth}
\textbf{\EWA{}, coverage 68\%}
\begin{verbatim}
## Basic Store
```typescript
import { createStore } from "zustand";
interface Counter {
  count: number;
  increment: () => void;
}
const useCounter = createStore<Counter>()((set) =>...
```
## Persist Middleware
\end{verbatim}
\end{minipage}

\medskip
\begin{minipage}[t]{0.485\linewidth}
\textbf{Original} (GSAP routing skill)
\begin{verbatim}
# GSAP Router
Routes to 4 specialized skills based on animation...
## Routing Protocol
1. **Classify** -- Identify animation type from u...
2. **Match** -- Apply signal matching rules below
3. **Combine** -- Production animations often nee...
4. **Load** -- Read matched SKILL.md files before...
\end{verbatim}
\end{minipage}\hfill
\begin{minipage}[t]{0.485\linewidth}
\textbf{\EWA{}, coverage 27\%}
\begin{verbatim}
# GS Router
Routes animations to specialized skills based on ...
## Routing Protocol
### Tier 1: Core Requirements (from signal)
1. **Class** -- Identify animation type from `ani...
2. **Combine** -- Apply combined SKILL routing be...
3. **Match** -- Production signals often need prod...
\end{verbatim}
\end{minipage}}
\caption{Original and recovered openings for two test skills with no near-copy in the training corpus (long lines truncated). Top: structure and
code are kept, one API name differs and a method is dropped. Bottom: the steps are reworded and a second tier is invented; the document is almost twice
as long as the original. At rank $64$ on recitation adapters, recovery is near-verbatim for hundreds of lines.}
\label{fig:examples}
\end{figure}

\paragraph{Self-authored agent tasks.} We also wrote one task per pilot skill ($41$ tasks with our own prompts, verifiers and reference solutions) and
ran the same Haiku agent: four tasks $8$ times per arm and $37$ tasks $3$ times. Because we wrote the verifiers, this is a weaker check than the benchmark
tasks. On $36$ tasks the agent passes with or without any skill, and the original and both reconstructions match it; two fail in every arm. On the three
tasks where the skill matters, the original passes $18$ of $19$ runs, \EWT{} $12$ and \EWA{} $11$, against $1$ with no skill; on threat triage \EWT{}
reproduces the original exactly ($8/8$ against $0/8$ with no skill) while \EWA{} fails. Over all $142$ runs per arm: no skill $117$, original $134$, \EWT{}
$130$, \EWA{} $125$ strict passes.

\paragraph{Benchmark task screening.} Of the $17$ pilot SkillsBench tasks we screened $13$ without a skill and kept six that the agent failed; two were
dropped after checking each with its reference solution. One needs a live download blocked in the sandbox; in the other, the expected value contradicts
the workbook's own rule (the filled cells of the same row are six-year averages, matching $6721.8$, $859.3$ and $1100.2$ exactly, while the verifier
expects a five-year average). With the skill given as a file in the sandbox instead of in the prompt ($3$ runs per task), the original and both
reconstructions passed $4$, $4$ and $5$ of $12$ runs, but the agent opened the skill in only about half of them.

\paragraph{Cost.} One document at a time the Executing Writer decodes about $8$ tokens per second ($69$ minutes for $37$ documents). The adapter layer
accepts a batch of adapters, so documents can be written together with each row using its own adapter; rows do not influence each other. With $8$
documents at a time the $37$ rank-$64$ adapters take $18.5$--$19.9$ minutes ($3.4$--$3.7\times$ faster) with unchanged coverage.

\section{Training Writers on Behavioural Adapters}
\label{app:ewb}

Section~\ref{sec:res-beh} showed that writers trained on recitation adapters do not read behavioural ones. Here the writer is trained on
behavioural adapters instead (\EWB{}). This is a smaller setting (Qwen3-4B base, judge-free metrics), so its numbers are not comparable with the main text.

\paragraph{Setup.} For each skill, the base model writes about $60$ requests and answers them with the document in its prompt; a rank-$16$ LoRA is fitted
to give those answers without the document. All $78$ fitted adapters lowered held-out answer loss by at least $10\%$ (median $69\%$). $54$ were used for
training and the rest for testing, together with $4$ skills made with the main paper's recipe. \EWB{} is the Executing Writer of Section~\ref{sec:method},
trained on $11$, $22$ and $45$ adapters. We compared it with training-free methods: generic and domain-targeted probes, vocabulary-lens words,
\emph{amplified answers} (sampling from $\log p_{\text{adapter}}+\alpha(\log p_{\text{adapter}}-\log p_{\text{base}})$, $\alpha=1$), best-of-$n$ sampling,
self-report and retrieval. Metrics are judge-free: \emph{behavioural fidelity} M1 is the share of the original document's help to the likelihood of
the prompted model's answers that a candidate provides ($1$ for the original, $0$ for none), and coverage counts the original's units that have an
embedding neighbour in the candidate (cosine $\ge0.757$, calibrated on documents that restate each other).

\begin{table}[h]
\centering
\caption{Judge-free recovery on held-out behavioural adapters (Qwen3-4B). Paper recipe: adapters made as in Section~\ref{sec:res-beh}.}
\label{tab:ewbres}
\small
\setlength{\tabcolsep}{4pt}
\begin{tabular}{llrrrrr}
\toprule
Set & Method & $n$ & M1, own & M1, wrong & Cov.\ own & Cov.\ wrong\\
\midrule
Held-out skills & Generic probes & 8 & $-0.14$ & $-0.05$ & $0.0$ & $0.0$\\
 & Vocabulary lens & 8 & $-0.04$ & $-0.09$ & $6.6$ & $0.0$\\
 & Amplified answers & 7 & \best{0.19} & $-0.07$ & $4.5$ & $0.0$\\
 & Plain request & 8 & $0.13$ & $0.02$ & $0.7$ & $0.0$\\
 & Best of $n$ & 7 & $0.11$ & $0.00$ & $5.6$ & $0.1$\\
 & Retrieval & 8 & $0.03$ & $-0.07$ & $1.9$ & $0.7$\\
 & \EWB{}, $45$ adapters & 8 & $0.11$ & $0.00$ & \best{7.3} & $1.2$\\
\midrule
Paper recipe & Amplified answers & 3 & $-0.01$ & $-0.04$ & $0.0$ & $0.0$\\
 & Plain request & 6 & $-0.00$ & $-0.09$ & $0.0$ & $0.0$\\
 & \EWB{}, $45$ adapters & 4 & $-0.02$ & $-0.05$ & $1.1$ & $0.0$\\
\bottomrule
\end{tabular}
\end{table}

\paragraph{Results.} The writer depends on the adapter: the correct adapter lowers validation loss by about $0.7$ nats per token against a wrong one,
against $0.001$ for the soft-prompt writer of Table~\ref{tab:gap}. But it overfits within $30$ of $120$ steps and memorises its training documents.
It has the best coverage on held-out skills ($7.3\%$, interval $[2.4,13.5]$, against $1.2\%$ for a wrong adapter; sign test $p=0.36$, $n=8$), far below
the $20\%$ we had set as success, and $1.1\%$ on the paper-recipe adapters (Table~\ref{tab:ewbres}). Its documents name the right area and miss the
specifics (for an architecture-diagram skill: ``AI Assistant for Code Review and Documentation''). Coverage rises from $0.0\%$ with $11$ training adapters
to $5.3\%$ with $45$, but only $2$ to $14$ documents were scored per size, so this is a direction, not a measurement. No training-free method beats the
wrong-adapter control on coverage by more than $4$ points; amplified answers are the exception on behavioural fidelity ($0.19$ against $-0.07$,
$p=0.011$, $n=10$), so they carry some adapter-specific behaviour without stating its rules. As a check of the stacking mechanism, the public Diff
Interpretation Tuning adapter~\cite{dit} loaded in the writer slot named a hidden topic from a weight difference in $11$ of $12$ cases at a reduced scale.
A bank of several hundred behavioural adapters with early stopping is the natural next step.

\section{Skills versus LoRA on Synthetic and Tool-Calling Tasks}
\label{app:fresh}

We compare a skill written by a Claude Haiku agent with a LoRA on six tasks with Qwen3.5-4B as the base model: four synthetic rule-following tasks with
programmatic scorers and known true skills (S1 house format, S2 fee calculation, S3 code table, S4 ticket triage) and two real tasks, R1 text-to-SQL and
R3 tool calling (the data of the public \texttt{doosra} adapters). Each task has $100$ development items, visible to the agent only through a scoring
command, and $200$ test items. LoRAs are rank $16$ on all projections, trained on $2{,}000$ examples; R1's LoRA gained only $+0.035$ on development and
was not read. Each skill is placed in the system prompt of the base model. Arms: \emph{P}, task text only; \emph{D}, task text and $32$, $128$ or $400$
training examples; \emph{A}, the adapter only, scored by how often base plus skill reproduces the adapter's output on development prompts;
\emph{A-wrong}, a different task's adapter; \emph{A+D}, adapter and data. One session per arm, so differences below about $0.05$ are within noise.

\begin{table}[h]
\centering
\caption{Test score ($200$ items) per task. Skills are applied to the frozen base model unless stated. ``--'': not defined or not run.}
\label{tab:fresh-main}
\small
\begin{tabular}{@{}l*{6}{r}@{}}
\toprule
Arm & S1 & S2 & S3 & S4 & R1 & R3\\
\midrule
Base model & 0.000 & 0.194 & 0.000 & 0.225 & 0.735 & 0.140\\
LoRA ($2{,}000$ examples) & 0.891 & \best{0.698} & \best{1.000} & \best{1.000} & \best{0.770} & \best{0.895}\\
True skill & 0.898 & 0.447 & 0.998 & 0.902 & -- & --\\
\midrule
Skill P (task text) & 0.909 & 0.404 & 0.000 & 0.860 & 0.730 & 0.190\\
Skill D ($400$ examples) & 0.955 & 0.472 & \best{1.000} & 0.993 & 0.665 & 0.245\\
Skill A (adapter only) & 0.927 & 0.454 & 0.970 & 0.935 & -- & 0.200\\
Skill A-wrong & 0.938 & 0.409 & 0.922 & 0.730 & -- & 0.145\\
Skill A+D & \best{0.964} & 0.515 & \best{1.000} & 0.955 & -- & 0.215\\
\midrule
Claude Haiku 5.5 with a skill & 1.000 & 0.958 & 1.000 & 0.983 & 0.730 & 0.630\\
\bottomrule
\end{tabular}
\end{table}

\paragraph{Results.} On the rule-like tasks S1, S3 and S4 a skill written from $400$ examples matches the LoRA trained on $2{,}000$, and with $32$ examples
the skill beats a LoRA trained on the same $32$ on five of the six tasks (most strongly on S3, $0.999$ against $0.302$; R3 is the exception). The LoRA wins where the
small model must execute something a prompt does not teach: on S2 (fee arithmetic) the LoRA scores $0.698$ and the best skill $0.531$, although Claude
Haiku with the same kind of skill scores $0.958$, so the document was right and the 4B model could not carry it out; on R3 (tool calling) the LoRA scores
$0.895$ and every skill $0.40$ or less. Skills written from the adapter alone recover $92$--$104\%$ of the LoRA's gain on S1, S3 and S4, but a skill from
another task's adapter recovers nearly as much on S1 and S3, because the agent infers those tasks from the description and the development feedback. With
repeated sessions (Table~\ref{tab:axdep}), the adapter clearly matters on S4. Giving the agent both adapter and data did not beat data alone.

\section{An Existing Adapter: Docker Run to Compose}
\label{app:dc}

To test reading on an adapter that was not built for this study, we used the \texttt{docker\_compose\_gen} LoRA that we had trained earlier for a
separate project and released in the \texttt{adaptive-minds-loras} collection (rank $32$,
Qwen2.5-7B-Instruct). The task is to convert a \texttt{docker run} command into the exact compose file that \texttt{docker compose} would print. Data:
$200$ test items over ten flag families ($20$ each), $100$ development and $400$ training items, all disjoint. A response is exact if its parsed YAML
equals the reference; the soft score is a field-level F1. Writers here are Claude Sonnet agents with the arms of Appendix~\ref{app:fresh}, plus
\emph{A-deep} (at least $25$ probes before the first score) and \emph{A+D-probe} (at least $20$ adapter probes before reading any example).

\begin{table}[h]
\centering
\caption{Docker-to-compose, $200$ test items: exact match and soft score. Rows with several sessions give the session mean.}
\label{tab:dc}
\small
\begin{tabular}{@{}lcc@{}}
\toprule
Arm & Exact & Soft\\
\midrule
Base model (Qwen2.5-7B-Instruct) & 0.000 & 0.186\\
LoRA \texttt{docker\_compose\_gen} & \best{0.595} & \best{0.885}\\
Wrong adapter (\texttt{dockerfile\_gen}) & 0.000 & 0.079\\
\midrule
Skill P (task text) & 0.000 & 0.340\\
Skill D (training data), 2 sessions & 0.313 & 0.819\\
Skill A (adapter only), 3 sessions & 0.263 & 0.771\\
Skill A-deep & 0.250 & 0.764\\
Skill A-wrong, 2 sessions & 0.000 & 0.278\\
Skill A+D-probe & 0.360 & 0.859\\
\midrule
$5$ / $20$ in-context examples (no agent) & 0.300 / 0.435 & 0.820 / 0.872\\
\midrule
LoRA $+$ skill P / A / A+D & 0.450 / 0.605 / \best{0.690} & 0.866 / 0.890 / \best{0.937}\\
\bottomrule
\end{tabular}
\end{table}

\paragraph{Results.} With no training data, after two to four probes of the adapter, the agent wrote skills that take the base model from $0.000$ to
$0.263$ exact on average ($0.235$--$0.300$) and recover about $84\%$ of the LoRA's soft gain. A skill from the wrong adapter and a skill from the task
text alone both stay at $0.000$, because the reference follows conventions the base model cannot guess (a fixed project name, a null default network,
a \texttt{command} list for tokens after the image). Probing harder did not help, but forcing probes before reading data gave the best skill ($0.360$).
Twenty in-context examples ($0.435$) beat every skill, and the LoRA ($0.595$) beats everything; a good skill on top of the LoRA adds up to $0.095$, and
a vague one costs up to $0.145$. Per flag family, adapter-only skills come close to the LoRA on logging ($0.38$ against $0.40$) but recover little on
numeric and structured fields (health checks $0.08$ against $0.70$). Renaming the service in $199$ test items left the LoRA unchanged ($0.593$), so it
had not memorised the test items. The one-session Haiku agent of Section~\ref{sec:ax} gave $0.212$ with the real adapter and $0.000$ with a wrong one.

\section{ALFWorld: Further Details}
\label{app:alf}

\paragraph{Environment and adapter.} Text version of ALFWorld (\texttt{alfworld} 0.4.2). Test: the $134$ \texttt{valid\_unseen} games, never shown to any
agent. Development: games from \texttt{valid\_seen}. The actor is Qwen2.5-7B-Instruct with plain chat prompts, greedy decoding and a step cap of $50$
(one-session study) or $52$ (reward loop and hidden rules). The LoRA (rank $16$, $\alpha=32$, all linear layers) was trained for one epoch on $1{,}477$
expert games and passed a pre-set gate of $+15$ points on development ($0.70$ against $0.10$).

\paragraph{Agent tools.} \texttt{task}; \texttt{rollout} (play a development game with the base or adapted model); \texttt{compare} (both models on the same
game, first divergence marked); \texttt{probe} (next action of both models at a written state); \texttt{counterfactual} and \texttt{logitdiff} (edit a
state, or show the tokens the adapter boosts); \texttt{try}; \texttt{score}; \texttt{examples} (demonstrations, for arms that have them). Across all
one-session ALFWorld runs the agents played $416$ rollouts, $93$ comparisons, $104$ probes and $138$ scores, and never used \texttt{counterfactual} or
\texttt{logitdiff}. Agents given the wrong adapter probed far more ($75$ probes against $13$), because comparing the two models showed them nothing.

\paragraph{Further one-session arms.} One retrieved expert game in the prompt gives $0.522$. Reading the adapter and then data gives $0.602$; skills written
from $50$ and from all demonstrations give $0.552$ and $0.527$, so more data did not help. The LoRA asked to write its own skill gives $0.149$, and an
automatic miner of base--adapter divergences followed by one writing call gives $0.256$ (sessions $0.57$, $0.06$, $0.13$). Giving the agent a mechanical
list of first divergences made the agent name rule A in every hidden-rule session and gave the steadiest win rate of any arm (session sd $0.026$).

\paragraph{Text on top of the LoRA.}
\begin{table}[h]
\centering
\caption{The ALFWorld LoRA with text added to its prompt ($134$ unseen games, step cap $52$).}
\label{tab:axlora}
\small
\begin{tabular}{lrr}
\toprule
Added to the LoRA & Success & Change\\
\midrule
Nothing & 0.813 & --\\
Short skill ($148$ words), system prompt & \best{0.858} & $+0.045$\\
Short skill, after the task text & 0.843 & $+0.030$\\
Three in-context games & 0.836 & $+0.023$\\
Three games $+$ full skill, system prompt & 0.821 & $+0.008$\\
Reward-loop skills, system prompt (mean of $12$) & 0.744--0.767 & $-0.05$ to $-0.07$\\
Full skill ($847$ words), system prompt & 0.716 & $-0.097$\\
Full skill, after the task text & 0.597 & $-0.216$\\
Three games $+$ full skill, after the task text & 0.425 & $-0.388$\\
\bottomrule
\end{tabular}
\end{table}

\paragraph{Adapter dependence against a task-hidden control.} Real adapter minus the task-hidden agent: ALFWorld $+0.085$ $[-0.081,+0.296]$; docker
$-0.028$ $[-0.137,+0.082]$ (the ``hidden'' docker arm still sees \texttt{docker run} in its prompts, so the task is not really hidden); S1 $-0.002$
$[-0.021,+0.017]$. This control was not run on S2, S4 and R3. On S3 one task-hidden session scored $0.63$, so S3 is dominated by session luck and we
draw no conclusion from it. R3 sessions hit the time limit on a shared server and have no valid result.

\paragraph{Portability details.} The one-session skill used on other actors is the session chosen on development success. A skill written from ten
demonstrations and chosen the same way gives $0.739$ on the home actor, $0.485$ on Llama-3.1-8B, $0.731$ on Qwen3.5-4B and $0.889$ on Claude Haiku ($18$
games, three per task type). On docker, transfer is uneven: the data-based skill gives exact match $0.96$ on Claude Haiku, $0.56$ on Qwen3.5-4B and $0.14$
on Llama, and the adapter-derived skill $0.44$, $0.11$ and $0.02$.

\subsection{Reward-loop protocol and audit}
\label{app:loop}

The $60$ games the writer sees and the $80$ selection games come from \texttt{valid\_seen} (shuffled, seed $0$); on them the base model scores $0.133$ and
$0.163$ and the LoRA $0.767$ and $0.813$. Each round is a fresh call of the writer in a persistent sandbox with a notes file as memory; up to three drafts
per round are played on the $60$ games and the best is played on the $80$. The linter rejected $43$ of $101$ draft attempts with the adapter and $60$ of
$115$ without, mostly for naming specific objects; every rejected draft could be fixed and resubmitted. One session lost two rounds to an unclear linter
message; dropping it moves the adapter-minus-no-adapter gap from $-2.0$ to $-1.9$ points. One further session produced no skill and is excluded. From
selection to test the mean score moved from $0.683$ to $0.653$ with the adapter and $0.698$ to $0.673$ without. No test win took fewer than three steps,
and a judge model read $20$ sampled wins per arm and found $39$ of $40$ legitimate; the exception ends as won after a heating step without a placing step,
using only valid commands, which we read as a scoring quirk of the environment.

\section{Hidden-Rule Experiments}
\label{app:rules}

\paragraph{Rules and scoring.} Rule A holds if the game is won and the action just before the winning action is exactly \texttt{inventory}. Rule B holds if
the game is won and the action right after the first successful \texttt{take} is exactly \texttt{look}. The original LoRA, the base model and the expert
games score $0\%$ on both rules; the expert games with the rule step inserted score $100\%$. LoRA-A and LoRA-B use the recipe of the ALFWorld LoRA with
the loss on the inserted step weighted three times, and passed a pre-set gate after one epoch (development rule success $87.5\%$ and $70.0\%$, base $0\%$).
For the adapter arms the agent's score is agreement with that adapter; no prompt, tool output or file the agent can read contains either rule.

\begin{table}[h]
\centering
\caption{Hidden-rule LoRAs on ALFWorld ($134$ unseen games). Rule columns give rule success: won and the rule followed.}
\label{tab:axrules}
\small
\setlength{\tabcolsep}{5pt}
\begin{tabular}{lrrrrc}
\toprule
Arm & Sessions & Win & Rule A & Rule B & Skill states its rule\\
\midrule
Base model & -- & 0.261 & 0.000 & 0.000 & --\\
LoRA-A & -- & 0.896 & \best{0.866} & 0.000 & --\\
LoRA-B & -- & 0.851 & 0.000 & \best{0.851} & --\\
\midrule
Agent reading LoRA-A & 6 & 0.604 & 0.005 & 0.000 & 5/6 (never rule B)\\
Agent reading LoRA-B & 6 & 0.565 & 0.000 & 0.053 & 5/6 (never rule A)\\
Agent reading LoRA-A with a divergence list & 6 & 0.636 & 0.001 & 0.000 & 6/6\\
Agent reading 10 rule-A demonstrations & 2 & 0.563 & 0.007 & 0.000 & 0/2\\
Agent with the base model only & 2 & 0.090 & 0.000 & 0.000 & 0/2\\
\midrule
Skill stating rule A outright & -- & 0.239 & 0.030 & 0.000 & yes\\
Skill stating rule B outright & -- & 0.209 & 0.007 & 0.194 & yes\\
\bottomrule
\end{tabular}
\end{table}

\paragraph{Contrasts.} Real minus decoy is $+0.005$ $[0.000,+0.016]$ for rule A and $+0.054$ $[+0.010,+0.119]$ for rule B, pooled $+0.029$
$[+0.006,+0.061]$, far below the pre-set $+15$ points. Counting every game, won or not, the LoRA-B actor says \texttt{look} right after its first take in
$99.3\%$ of games, the actor with the outright skill in $67.9\%$, and actors with agent-written skills in $10.3\%$. The actor with the outright rule-A skill
says \texttt{inventory} somewhere in $84.3\%$ of games but just before the last step in only $7.5\%$. The actor can be made to say the word, but not at
the right moment.

\paragraph{Rules the actor can follow.} We tested six candidate rules by appending each to a competent skill and measuring compliance on development
games: \texttt{look} first ($25\%$), \texttt{inventory} first ($45\%$), examine the taken object right after the first take ($53\%$), \texttt{inventory}
right after the first take ($65\%$), \texttt{look} after the first move ($20\%$), and \texttt{inventory} as the second action ($0\%$). The two above $50\%$ were
used. LoRAs trained for them win $78\%$ of development games and follow their rule in $78\%$, and are scored under each other's rule as decoys. With
three reward-loop sessions per rule (three rounds each), test rule success was $0.09$, $0.41$, $0.25$ for \texttt{inventory} after the first take and $0.00$,
$0.22$, $0.00$ for examining the object, against $0.00$ for the decoy rule in all six sessions; win rates were $0.59$--$0.74$.

\section{Reading Weights}
\label{app:weights}

\paragraph{Lenses.} The weight lens takes the top singular directions of each LoRA update in the attention-output and MLP-down projections and reads
them through the final norm and output embedding, with and without centring over the three ALFWorld LoRAs. For every LoRA, most ALFWorld command words
reach rank $1$ in some layer, direction and sign, so the best rank is not specific. The activation lens averages, over $200$ development states, the
difference between the adapted and base model's last-token hidden state at each layer, subtracts the mean of this difference over the three LoRAs, and
reads it through the output embedding.

\paragraph{Ablations.} Keeping the update of LoRA-A only in some layers or modules, on $60$ development games (win / rule A): full $0.90$ / $0.90$;
layers $0$--$8$ $0.48$ / $0.00$; layers $9$--$18$ $0.83$ / $0.52$; layers $19$--$27$ $0.50$ / $0.48$; MLP only $0.92$ / $0.92$; attention only $0.77$ /
$0.07$. For the ALFWorld LoRA, layers $9$--$18$ alone give $0.93$ and MLP only $0.97$ on development, above the full adapter's $0.77$; differences of a few
games are within noise.


\begin{thebibliography}{99}
\bibitem{anthropicskills} Anthropic.
\newblock Equipping agents with Agent Skills.
\newblock Anthropic Engineering blog, 2025. https://www.anthropic.com/engineering/equipping-agents-for-the-real-world-with-agent-skills

\bibitem{skillsbench} X.~Li et al.
\newblock SkillsBench: Benchmarking How Well Agent Skills Work Across Diverse Tasks.
\newblock 2026. arXiv:2602.12670.

\bibitem{lora} E.~J.~Hu et al.
\newblock LoRA: Low-Rank Adaptation of Large Language Models.
\newblock \emph{ICLR}, 2022. arXiv:2106.09685.

\bibitem{t2l} R.~Charakorn et al.
\newblock Text-to-LoRA: Instant Transformer Adaption.
\newblock \emph{ICML}, 2025. arXiv:2506.06105.

\bibitem{d2l} R.~Charakorn, E.~Cetin, S.~Uesaka and R.~T.~Lange.
\newblock Doc-to-LoRA: Learning to Instantly Internalize Contexts.
\newblock \emph{ICML}, 2026. arXiv:2602.15902.

\bibitem{shine} Y.~Liu, X.~Wang, Y.~Mao, Y.~Gelberg, H.~Maron and M.~Zhang.
\newblock SHINE: A Scalable In-Context Hypernetwork for Mapping Context to LoRA in a Single Pass.
\newblock \emph{ICML}, 2026. arXiv:2602.06358.

\bibitem{alfworld} M.~Shridhar, X.~Yuan, M.-A.~C\^ot\'e, Y.~Bisk, A.~Trischler and M.~Hausknecht.
\newblock ALFWorld: Aligning Text and Embodied Environments for Interactive Learning.
\newblock In \emph{ICLR}, 2021.

\bibitem{maia} T.~R.~Shaham, S.~Schwettmann, F.~Wang, A.~Rajaram, E.~Hernandez, J.~Andreas and A.~Torralba.
\newblock A Multimodal Automated Interpretability Agent.
\newblock In \emph{ICML}, 2024.

\bibitem{latentskill} A.~Yu et al.
\newblock LatentSkill: From In-Context Textual Skills to In-Weight Latent Skills for LLM Agents.
\newblock 2026. arXiv:2606.06087.

\bibitem{paramskills} X.~Zhao et al.
\newblock Parametric Skills.
\newblock 2026. arXiv:2606.30015.

\bibitem{slora} Y.~Sheng et al.
\newblock S-LoRA: Serving Thousands of Concurrent LoRA Adapters.
\newblock \emph{MLSys}, 2024. arXiv:2311.03285.

\bibitem{w2t} X.~Han, F.~Neri, Z.~Jiang, F.~Wu, Y.~Ye, L.~Yin and Z.~Wang.
\newblock W2T: LoRA Weights Already Know What They Can Do.
\newblock 2026. arXiv:2603.15990.

\bibitem{loracles} C.~De~Schamphelaere, J.~Bauer, N.~Nanda, E.~Ong.
\newblock LoRAcles: Self-Supervised Weight-Space Interpretability at Scale.
\newblock \emph{NeurIPS}, 2026.

\bibitem{dit} A.~Goel, Y.~Kim, N.~Shavit and T.~T.~Wang.
\newblock Learning to Interpret Weight Differences in Language Models (Diff Interpretation Tuning).
\newblock \emph{ICLR}, 2026. arXiv:2510.05092.

\bibitem{weightwatch} Z.~Zhong and A.~Raghunathan.
\newblock Watch the Weights: Unsupervised Monitoring and Control of Fine-tuned LLMs.
\newblock \emph{ICLR}, 2026. arXiv:2508.00161.

\bibitem{spectral} R.~Paul.
\newblock Spectral Geometry of LoRA Adapters Encodes Training Objective and Predicts Harmful Compliance.
\newblock 2026. arXiv:2604.08844.

\bibitem{voyager} G.~Wang et al.
\newblock Voyager: An Open-Ended Embodied Agent with Large Language Models.
\newblock \emph{TMLR}, 2024. arXiv:2305.16291.

\bibitem{react} S.~Yao et al.
\newblock ReAct: Synergizing Reasoning and Acting in Language Models.
\newblock \emph{ICLR}, 2023. arXiv:2210.03629.

\bibitem{sweagent} J.~Yang et al.
\newblock SWE-agent: Agent-Computer Interfaces Enable Automated Software Engineering.
\newblock \emph{NeurIPS}, 2024. arXiv:2405.15793.

\bibitem{houlsby} N.~Houlsby et al.
\newblock Parameter-Efficient Transfer Learning for NLP.
\newblock \emph{ICML}, 2019. arXiv:1902.00751.

\bibitem{prefix} X.~L.~Li, P.~Liang.
\newblock Prefix-Tuning: Optimizing Continuous Prompts for Generation.
\newblock \emph{ACL}, 2021. arXiv:2101.00190.

\bibitem{intrinsic} A.~Aghajanyan, L.~Zettlemoyer, S.~Gupta.
\newblock Intrinsic Dimensionality Explains the Effectiveness of Language Model Fine-Tuning.
\newblock \emph{ACL}, 2021. arXiv:2012.13255.

\bibitem{lorabiderman} D.~Biderman et al.
\newblock LoRA Learns Less and Forgets Less.
\newblock 2024. arXiv:2405.09673.

\bibitem{hypernet} D.~Ha, A.~Dai, Q.~V.~Le.
\newblock HyperNetworks.
\newblock \emph{ICLR}, 2017. arXiv:1609.09106.

\bibitem{hypter} Q.~Ye, X.~Ren.
\newblock Learning to Generate Task-Specific Adapters from Task Description (HYPTER).
\newblock \emph{ACL}, 2021. arXiv:2101.00420.

\bibitem{askell} A.~Askell et al.
\newblock A General Language Assistant as a Laboratory for Alignment.
\newblock 2021. arXiv:2112.00861.

\bibitem{snell} C.~Snell, D.~Klein, R.~Zhong.
\newblock Learning by Distilling Context.
\newblock 2022. arXiv:2209.15189.

\bibitem{gist} J.~Mu, X.~L.~Li, N.~Goodman.
\newblock Learning to Compress Prompts with Gist Tokens.
\newblock \emph{NeurIPS}, 2023. arXiv:2304.08467.

\bibitem{cartridges} S.~Eyuboglu et al.
\newblock Cartridges: Lightweight and General-Purpose Long Context Representations via Self-Study.
\newblock 2025. arXiv:2506.06266.

\bibitem{lorahub} C.~Huang et al.
\newblock LoraHub: Efficient Cross-Task Generalization via Dynamic LoRA Composition.
\newblock 2023. arXiv:2307.13269.

\bibitem{taskarith} G.~Ilharco et al.
\newblock Editing Models with Task Arithmetic.
\newblock \emph{ICLR}, 2023. arXiv:2212.04089.

\bibitem{lol} T.~Putterman et al.
\newblock Learning on LoRAs: GL-Equivariant Processing of Low-Rank Weight Spaces for Large Finetuned Models.
\newblock 2024. arXiv:2410.04207.

\bibitem{hyperrep} K.~Sch\"urholt et al.
\newblock Hyper-Representations as Generative Models: Sampling Unseen Neural Network Weights.
\newblock 2022. arXiv:2209.14733.

\bibitem{nndiff} K.~Wang et al.
\newblock Neural Network Diffusion.
\newblock 2024. arXiv:2402.13144.

\bibitem{logitlens} nostalgebraist.
\newblock Interpreting GPT: the logit lens.
\newblock LessWrong, 2020. https://www.lesswrong.com/posts/AcKRB8wDpdaN6v6ru

\bibitem{tunedlens} N.~Belrose et al.
\newblock Eliciting Latent Predictions from Transformers with the Tuned Lens.
\newblock 2023. arXiv:2303.08112.

\bibitem{patchscopes} A.~Ghandeharioun et al.
\newblock Patchscopes: A Unifying Framework for Inspecting Hidden Representations of Language Models.
\newblock \emph{ICML}, 2024. arXiv:2401.06102.

\bibitem{selfie} H.~Chen et al.
\newblock SelfIE: Self-Interpretation of Large Language Model Embeddings.
\newblock \emph{ICML}, 2024. arXiv:2403.10949.

\bibitem{betley} J.~Betley et al.
\newblock Tell me about yourself: LLMs are aware of their learned behaviors.
\newblock 2025. arXiv:2501.11120.

\bibitem{carlini} N.~Carlini et al.
\newblock Extracting Training Data from Large Language Models.
\newblock \emph{USENIX Security}, 2021. arXiv:2012.07805.

\bibitem{nasr} M.~Nasr et al.
\newblock Scalable Extraction of Training Data from (Production) Language Models.
\newblock 2023. arXiv:2311.17035.

\bibitem{vec2text} J.~X.~Morris et al.
\newblock Text Embeddings Reveal (Almost) As Much As Text.
\newblock \emph{EMNLP}, 2023. arXiv:2310.06816.

\bibitem{lminversion} J.~X.~Morris et al.
\newblock Language Model Inversion.
\newblock \emph{ICLR}, 2024. arXiv:2311.13647.

\bibitem{infonce} A.~van~den~Oord, Y.~Li, O.~Vinyals.
\newblock Representation Learning with Contrastive Predictive Coding.
\newblock 2018. arXiv:1807.03748.

\bibitem{clip} A.~Radford et al.
\newblock Learning Transferable Visual Models From Natural Language Supervision.
\newblock \emph{ICML}, 2021. arXiv:2103.00020.

\bibitem{qwen3emb} Y.~Zhang et al.
\newblock Qwen3 Embedding: Advancing Text Embedding and Reranking Through Foundation Models.
\newblock 2025. arXiv:2506.05176.

\bibitem{mtbench} L.~Zheng et al.
\newblock Judging LLM-as-a-Judge with MT-Bench and Chatbot Arena.
\newblock \emph{NeurIPS Datasets and Benchmarks}, 2023. arXiv:2306.05685.

\bibitem{vllm} W.~Kwon et al.
\newblock Efficient Memory Management for Large Language Model Serving with PagedAttention.
\newblock \emph{SOSP}, 2023. arXiv:2309.06180.

\bibitem{qwen3} A.~Yang et al.
\newblock Qwen3 Technical Report.
\newblock 2025. arXiv:2505.09388.
\bibitem{ape} Y.~Zhou et al.
\newblock Large Language Models Are Human-Level Prompt Engineers.
\newblock \emph{ICLR}, 2023. arXiv:2211.01910.

\bibitem{opro} C.~Yang et al.
\newblock Large Language Models as Optimizers.
\newblock \emph{ICLR}, 2024. arXiv:2309.03409.

\bibitem{dspy} O.~Khattab et al.
\newblock DSPy: Compiling Declarative Language Model Calls into Self-Improving Pipelines.
\newblock \emph{ICLR}, 2024. arXiv:2310.03714.

\bibitem{textgrad} M.~Yuksekgonul et al.
\newblock Optimizing generative AI by backpropagating language model feedback (TextGrad).
\newblock \emph{Nature}, 2025. arXiv:2406.07496.

\bibitem{gepa} L.~A.~Agrawal et al.
\newblock GEPA: Reflective Prompt Evolution Can Outperform Reinforcement Learning.
\newblock 2025. arXiv:2507.19457.

\bibitem{reflexion} N.~Shinn et al.
\newblock Reflexion: Language Agents with Verbal Reinforcement Learning.
\newblock \emph{NeurIPS}, 2023. arXiv:2303.11366.

\bibitem{expel} A.~Zhao et al.
\newblock ExpeL: LLM Agents Are Experiential Learners.
\newblock \emph{AAAI}, 2024. arXiv:2308.10144.

\bibitem{awm} Z.~Z.~Wang et al.
\newblock Agent Workflow Memory.
\newblock 2024. arXiv:2409.07429.

\bibitem{auditing} S.~Marks et al.
\newblock Auditing Language Models for Hidden Objectives.
\newblock 2025. arXiv:2503.10965.

\bibitem{qwen25} Qwen Team.
\newblock Qwen2.5 Technical Report.
\newblock 2024. arXiv:2412.15115.

\bibitem{llama3} A.~Grattafiori et al.
\newblock The Llama 3 Herd of Models.
\newblock 2024. arXiv:2407.21783.
\end{thebibliography}
\end{document}